\documentclass[11pt]{article}
\usepackage{enumerate}
\usepackage{natbib}
\usepackage{amsmath,amsthm,amssymb}
\usepackage{lmodern}
\usepackage{iftex}
\usepackage{xcolor}
\usepackage[utf8]{inputenc}

\usepackage{authblk}
\usepackage{graphicx,psfrag,epsf}
\usepackage{epstopdf}
\usepackage{epsfig}
\usepackage[toc]{appendix}
\usepackage{bbm}
\usepackage{enumitem}
\usepackage{pdflscape}
\usepackage{afterpage}

\usepackage{url} 
\usepackage[ruled,vlined]{algorithm2e}
\usepackage{algorithmicx,algpseudocode}
\usepackage{setspace}
\usepackage{multirow}  
\usepackage{natbib}
\usepackage{soul} 
\usepackage{rotating}

\SetKwInput{KwInput}{Input}                
\SetKwInput{KwOutput}{Output}

\setcitestyle{square,numbers}

\usepackage{datatool}
\usepackage{booktabs}

\RequirePackage[hyperindex,breaklinks,colorlinks,citecolor=blue,urlcolor=black]{hyperref} 

\theoremstyle{plain}
\newcommand{\blind}{0}

\newcommand{\textBF}[1]{%
	\pdfliteral direct {2 Tr 0.5 w} 
	#1%
	\pdfliteral direct {0 Tr 0 w}%
}

\begin{document}

\def\spacingset#1{\renewcommand{\baselinestretch}%
{#1}\small\normalsize} \spacingset{1}

\if0\blind
{
  \title{Unsupervised outlier detection using random subspace and subsampling ensembles of Dirichlet process mixtures\thanks{Dongwook Kim and Juyeon Park contributed equally to this work.}}
  \author[1]{Dongwook Kim}
  \author[2]{Juyeon Park}
  \author[3,4]{Hee Cheol Chung}
  \author[5,6]{Seonghyun Jeong\thanks{Corresponding author: sjeong@yonsei.ac.kr}} 
  \affil[1]{Nexon, Seongnam-si, Gyeonggi-do, Korea}
   \affil[2]{Danggeun Market, Seoul, Korea}
  \affil[3]{Department of Mathematics and Statistics, University of North Carolina at Charlotte, Charlotte, North Carolina, USA}
    \affil[4]{School of Data Science, University of North Carolina at Charlotte, Charlotte, North Carolina, USA}
     \affil[5]{Department of Statistics and Data Science, Yonsei University, Seoul, Korea}
  \affil[6]{Department of Applied Statistics, Yonsei University, Seoul, Korea}
  \maketitle
} \fi

\begin{abstract}

Probabilistic mixture models are recognized as effective tools for unsupervised outlier detection owing to their interpretability and global characteristics. Among these, Dirichlet process mixture models stand out as a strong alternative to conventional finite mixture models for both clustering and outlier detection tasks. Unlike finite mixture models, Dirichlet process mixtures are infinite mixture models that automatically determine the number of mixture components based on the data. 
Despite their advantages, the adoption of Dirichlet process mixture models for unsupervised outlier detection has been limited by challenges related to computational inefficiency and sensitivity to outliers in the construction of outlier detectors.
Additionally, Dirichlet process Gaussian mixtures struggle to effectively model non-Gaussian data with discrete or binary features.
To address these challenges, we propose a novel outlier detection method that utilizes ensembles of Dirichlet process Gaussian mixtures. This unsupervised algorithm employs random subspace and subsampling ensembles to ensure efficient computation and improve the robustness of the outlier detector. The ensemble approach further improves the suitability of the proposed method for detecting outliers in non-Gaussian data.
Furthermore, our method uses variational inference for Dirichlet process mixtures, which ensures both efficient and rapid computation. Empirical analyses using benchmark datasets demonstrate that our method outperforms existing approaches in unsupervised outlier detection.

\end{abstract}

\noindent {\it Keywords: Anomaly detection, Gaussian mixture models, outlier ensembles, random projection, variational inference}

\spacingset{1.0}



\section{Introduction}
\label{sec:intro}

The era of big data has resulted in an overwhelming  influx of information, including both relevant and irrelevant observations. As a result, identifying and detecting these irrelevant portions of data, known as outliers, has become increasingly important, as they can obscure the dominant patterns and characteristics of the overall dataset. 
Outlier detection has been explored across  various research communities, including statistics, computer science, and information theory. Typically, outliers are instances that deviate significantly from the majority of the dataset.  The fundamental goal of outlier detection is to identify a model that effectively distinguishes these nonconforming instances as outliers. However, defining what constitutes normal heavily depends on the specific data domain, and this critical information is often not available beforehand.

To address this challenge, various unsupervised methods have been proposed, including both probabilistic and non-probabilistic approaches \citep{yang2021mean,li2022ecod,goodge2022lunar,xu2023deep,arias2023aida,mensi2023detecting,xu2023fascinating,tu2024weighted}. Probabilistic methods offer a notable advantage owing to their interpretability, which stems  from their solid statistical foundations \citep{aggarwal2017outlier}. These approaches provide clear  insights into the degree of anomalies by assigning probabilities or likelihood scores to individual data points. Additionally, their model specification allows for the quantification of uncertainty in the measured degree of anomalies.

Within probabilistic methods, mixture models have gained significant attention for modeling heterogeneous populations \citep{2bishop2006pattern,5laxhammar2009anomaly,6li2016anomaly,7veracini2009fully,8bahrololum2008anomaly,mclachlan2019finite}.
From a Bayesian perspective, Dirichlet process (DP) mixtures have become a prominent framework for probabilistic mixture models. They address the limitations of finite mixture models by allowing for an infinite number of mixture components \citep{15sethuraman1994constructive,gelman2013bayesian}. This feature enables the model to determine the optimal number of mixture components in a data-driven manner, offering greater flexibility in capturing the underlying data structure. DP mixtures have been used for outlier detection in various fields \citep{56shotwell2011bayesian,kaltsa2018multiple,arisoy2021nonparametric}.
However, the training of mixture parameters is often significantly  influenced by potential outliers, which can introduce substantial bias into the parameter estimates \citep{garcia2011tclust,punzo2016parsimonious}. This bias must be carefully managed when applying mixture models to outlier detection tasks.
Additionally, estimating  clustering memberships in mixture models can be computationally intensive, which may be a major bottleneck, making mixture-based outlier detection methods considerably slower compared to non-probabilistic approaches.

In this study, we propose a novel outlier detection method based on the DP mixture framework. To address the issues associated with DP mixture models in outlier detection, our method incorporates two key concepts: variational inference and outlier ensemble analysis. First, variational inference aims to find the distribution that best approximates the  posterior distribution by minimizing the Kullback-Leibler (KL) divergence \citep{jordan1999introduction}. As a computationally faster alternative to Markov chain Monte Carlo (MCMC) methods, variational inference effectively mitigates the computational inefficiency of DP mixture models.
We build on the variational algorithm for DP mixture models developed by \cite{17blei2006variational}. For a detailed discussion on variational inference for DP mixture models, refer to Section~\ref{sec:dp}.
Additionally, the concept of outlier ensembles is employed to enhance outlier detection performance by leveraging the collective wisdom of multiple weak learners. By aggregating the results from various base detectors--each specializing in different aspects of outliers--outlier ensembles can improve robustness and potentially reduce  computational costs. 
Ensemble analysis has a well-established history in classification \citep{32breiman2001random} and clustering \citep{34strehl2002cluster}, and has more recently been applied to outlier detection \citep{26kriegel2009outlier, 27lazarevic2005feature, 28keller2012hics, 30liu2008isolation}. \citet{35aggarwal2015} justifies the use of ensemble analysis for outlier detection in terms of the bias-variance tradeoff. Our approach utilizes two types of ensembles: subspace ensembles, which reduce the dimensionality of the feature space, and subsampling ensembles, which  reduce the number of instances. Each type has distinct advantages, detailed in Section~\ref{sec3}. We demonstrate that ensemble analysis allows non-Gaussian data to be effectively modeled by Gaussian mixture models, significantly reducing computation time without compromising detection accuracy.
By combining variational inference with outlier ensembles, our method--based on DP mixture models--achieves exceptional detection accuracy on benchmark datasets. This integration results in a robust and highly accurate outlier detection approach. The Python module for the proposed method is available at \url{https://github.com/juyeon999/OEDPM}. Key aspects of the proposed method are summarized as follows.

\begin{itemize}
	\item \emph{Interpretation.} The proposed method builds on the DP mixture framework for outlier detection, offering natural insights into the degree of anomalies through likelihood values.
	\item \emph{Automatic model determination.} Finite Gaussian mixtures are sensitive to the choice of the number of mixture components, requiring post-processing for model selection. In contrast, DP mixtures are infinite mixture models that determine the number of actual mixture components in a data-driven manner.
	\item \emph{Fast computation.} Mixture models, including finite Gaussian mixtures and DP mixtures, are typically computationally expensive. We enhance computational efficiency by employing variational inference and ensemble analysis.
	\item \emph{Modeling of non-Gaussian data.} Although DP mixtures use Gaussian distributions, many real datasets deviate from this assumption. The proposed method can effectively handle non-Gaussian data, including discrete or binary data, through subspace ensembles with random projections.
	\item \emph{Outlier-free training of the detector.} The performance of a detection model can be compromised if the training procedure is affected by outliers. Our methods aim to eliminate this issue by pruning irrelevant mixture components, thereby reducing the influence of outliers.
	\item \emph{Python module.} The Python module for the proposed method is readily available.
\end{itemize}

The remainder of this paper is organized as follows. Section~\ref{sec:rw} reviews the literature on mixture models and outlier ensembles for outlier detection tasks. 
Section~\ref{sec2} presents the foundational elements of the proposed method, including the variational algorithm for DP mixture models and comprehensive details on outlier ensembles. Section~\ref{sec4} provides specific details of the proposed method for unsupervised outlier detection. Section~\ref{sec:data} presents numerical analyses using real benchmark datasets. Finally, Section~\ref{sec7} concludes the study with a discussion summarizing the key findings and their implications.

\section{Related works}
\label{sec:rw}

\subsection{Mixture models for outlier detection}
\label{sec:rw.mm}

Gaussian mixture models (GMMs) have proven effective for various outlier detection tasks across different domains, including maritime \citep{5laxhammar2009anomaly}, aviation \citep{6li2016anomaly}, hyperspectral imagery \citep{7veracini2009fully}, and security systems \citep{8bahrololum2008anomaly}. A finite GMM assumes that each instance is generated from a mixture of multivariate Gaussian distributions \citep{mclachlan2019finite}.
Within this framework, the likelihood can naturally serve as an outlier score, as anomalous points exhibit significantly small likelihood values \citep{aggarwal2017outlier}. One advantage of this approach is that the resulting outlier score reflects the global characteristics of the entire dataset rather than just local properties. Furthermore, the outlier score derived from a GMM is closely related to the Mahalanobis distance, which accounts for inter-attribute correlations by dividing each coordinate value by the standard deviation in each direction. Consequently, the outlier score accounts for the relative scales of each dimension \citep{aggarwal2017outlier}.

Choosing the appropriate number of mixture components in a GMM is crucial, as it significantly affects the model's overall performance.
The conventional method involves conducting a sensitivity analysis using model selection criteria such as the Bayesian information criterion (BIC) \citep{5laxhammar2009anomaly, 6li2016anomaly, 8bahrololum2008anomaly}. However,  determining the optimal number of mixture components is challenging in outlier detection tasks as the presence of outliers can influence the selection procedure. Several attempts have been made to address this issue.
For instance, \citet{garcia2011tclust} introduced a method allowing a fraction of data points to belong to extraneous distributions, which are excluded during GMM training. 
\citet{punzo2016parsimonious} considered a contaminated mixture model by replacing the Gaussian components of GMMs with contaminated Gaussian distributions, defined as mixtures of two Gaussian components for inliers and outliers. Despite these attempts, using model selection criteria like BIC has the disadvantage of requiring post-comparison of GMM fits across various numbers of components.

Another appealing approach is to automate the search for the optimal number of mixture components within the inferential procedure in a data-driven manner.
One effective method to achieve this is by incorporating a DP prior within the Bayesian framework, resulting in a procedure known as the DP mixture model \citep{gelman2013bayesian}.
\citet{56shotwell2011bayesian} first employed DP mixtures for outlier detection, treating it as a clustering task in general scenarios. Since then, this method has been used in various areas, including image analysis \citep{arisoy2021nonparametric}, video processing \citep{kaltsa2018multiple}, and human dynamics \citep{fuse2017statistical}.
Similar to GMMs, DP mixtures face challenges due to the unsupervised nature of outliers, as the overall training procedure relies on the full dataset, including outliers. However, to the best of our knowledge, no attempts have been made to address this issue within the DP mixture framework.

\subsection{Ensemble analysis for outlier detection}

Real-world datasets present practical challenges in outlier detection due to their typically large number of features and instances. Additional dimensions do not necessarily provide more information about the outlying nature of specific data points. As noted in previous studies \citep{chung2021subspace}, data points in high-dimensional spaces often converge towards the vertices of a simplex, resulting in similar pairwise distances among instances. This phenomenon makes distance-based detection models ineffective at distinguishing outliers from normal instances. 
Additionally, having a large number of instances does not necessarily enhance the identification of abnormal instances \citep{muhr2022little}. For example, \citet{30liu2008isolation} found that a large number of instances may lead to masking and swamping effects. The masking effect occurs when extreme instances cause other extreme instances to appear normal, while the swamping effect occurs when densely clustered normal instances are mistakenly flagged as outliers. Consequently, to improve the robustness of outlier detectors, reducing the number of features and instances is often recommended \citep{35aggarwal2015}.

To address the challenge of high-dimensional features, subspace outlier detection methods have been proposed to identify informative subspaces where outlying points exhibit significant deviations from normal behavior \citep{26kriegel2009outlier, 27lazarevic2005feature, 28keller2012hics,30liu2008isolation}. However, exploring subspaces directly can be computationally expensive and sometimes infeasible owing to the exponential increase in the number of potential dimensions. A practical and effective approach is to form an ensemble of weakly relevant  subspaces using random mechanisms such as random projection \citep{bingham2001random}. Ensemble-based analysis has demonstrated significant advantages in high-dimensional outlier detection owing to its flexibility and robustness \citep{aggarwal2017outlier}. This approach, often referred to as rotated bagging \citep{35aggarwal2015}, aggregates results from all ensemble subspaces, which are derived by applying base detectors in lower-dimensional spaces. Unlike other outlier detection methods, the subspace ensemble approach has not been widely adopted for GMM-based outlier detection models, with one notable exception being its application in cyberattack detection \citep{an2022ensemble}.

On one hand, the subsampling outlier ensemble method addresses the challenge of managing a large number of instances by randomly selecting instances from the dataset without replacement. This process generates weakly relevant training data for each component of the ensemble. In this context, subsampling creates a collection of subsamples that act as ensemble components. This concept is related to bagging \citep{breiman1996bagging}, though bagging relies on bootstrap samples generated by sampling with replacement.
The use of subsampling in outlier ensembles was initially prominent with the isolation forest \citep{30liu2008isolation}, where it contributed to improved computational efficiency.
Additionally, subsampling has proven effective in enhancing outlier detection accuracy in proximity-based methods, such as local outlier factors and nearest neighbors \citep{zimek2013subsampling,35aggarwal2015}. Despite the promising attributes of the subsampling ensemble method, further investigation is needed to determine its effectiveness in improving GMM-based outlier detection models.

\section{Fundamentals of the proposed method}
\label{sec2}

While detailed information is provided in Section~\ref{sec4}, a brief outline of the proposed outlier detection method is as follows.
Let $\mathbf D\in\mathbb{R}^{N\times p}$ be the training dataset. For $m=1,\dots,M$, let $\mathbf X_m\in\mathbb R^{n_m\times d_m}$ denote datasets reduced from $\mathbf D$, where $n_m\le N$ and $d_m\le p$. For a mixture model with a specified density $p_m:\mathbb R^{d_m}\rightarrow[0,\infty)$ for reduced instances, each $\mathbf X_m$ is used to train a fitted density $\hat p_m:\mathbb R^{d_m}\rightarrow[0,\infty)$ of $p_m$. Consider $\mathbf d^{\text{new}}\in \mathbb R^{p}$ as a new (test) instance, and $\mathbf x_m^{\text{new}}\in \mathbb R^{d_m}$, $m=1,\dots,M$, as reduced instances generated by the same process as the training dataset. An outlier score for $\mathbf d^{\text{new}}$ is obtained based on the likelihood values $\hat p(\mathbf x_1^{\text{new}}),\dots, \hat p(\mathbf x_M^{\text{new}})$.

A complete description of the method involves specifying a model with density $p_m$ for reduced instances and detailing the data reduction process that generates each $\mathbf X_m$ from $\mathbf D$. Our approach uses the DP mixture framework for modeling and employs subspace and subsampling ensembles for data reduction. We provide a detailed explanation of the DP mixture framework in Section~\ref{sec:dp} and discuss  the ensemble analysis for the proposed method in Section~\ref{sec3}.

\subsection{Dirichlet process mixtures for the proposed method}

\label{sec:dp}

In this section, we describe the DP mixture model that specifies the density $p_m$ for reduced data of dimension $d_m$. We also discuss variational inference used to construct the fitted density $\hat p_m$ using each $\mathbf X_m$. Detailed information on $\hat p_m$, along with a pruning procedure to remove the effects of outliers in training, is provided in Section~\ref{sec4}. Since the procedures are consistent across all ensemble components $m=1,\dots, M$, we omit the subscript $m$ throughout Section~\ref{sec:dp}. Consequently, we use $\mathbf X =[\mathbf{x}_1, \dots, \mathbf{x}_n ]^T\in\mathbb R^{n\times d}$ to denote a reduced data matrix of dimension $n\times d$.

\subsubsection{Dirichlet process mixture models}
\label{sec:dpmm}

A finite GMM with $K$ mixture components is defined as a weighted sum of multivariate Gaussian distributions. For a $d$-dimensional instance $\mathbf x_i\in\mathbb R^d$, its mixture density is given by $\sum_{k=1}^K \pi_k  \varphi_d(\cdot\,;\boldsymbol\mu_k, \boldsymbol\Sigma_k )$, where $\varphi_d(\cdot\,;\boldsymbol\mu, \boldsymbol\Sigma)$ represents the $d$-dimensional Gaussian density with mean $\boldsymbol \mu$ and covariance matrix $\boldsymbol\Sigma$, and $\pi_k$ are the mixture weights such that $\pi_k>0$ and $\sum_{k=1}^K \pi_k = 1$.  
This mixture density defines the likelihood, which can be used to determine outlier scores for specific instances.

As previously mentioned, choosing an appropriate value for $K$ is crucial and can pose challenges in outlier detection. An alternative approach is to use DP mixture models.
The DP is a stochastic process that serves as a prior distribution over the space of probability measures. Consider a random probability measure $P$ defined over $(\Omega, \mathcal B)$, where $\Omega$ represents the sample space and $\mathcal B$ is the Borel $\sigma$-field encompassing all possible subsets of $\Omega$. With a parametric base distribution $P_0$ and a concentration parameter $\alpha > 0$, $P$ follows a DP if, for any finite partition ${B_1, \dots, B_H }$ of $\Omega$ with any finite $H$, the distribution of $\big( P{(B_1)}, \dots, P({B_H}) \big)$ is a Dirichlet distribution:
\begin{align}
	\big( P{(B_1)}, \dots, P({B_H}) \big ) \sim \text{Dirichlet}\big( \alpha P_0(B_1), \dots, \alpha P_0(B_H) \big).
\end{align}
This is denoted as $P \sim \text{DP}(\alpha, P_0)$. The concentration parameter $\alpha$ controls how $P$ concentrates around the base distribution $P_0$. A common default choice is $\alpha = 1$ \citep{gelman2013bayesian}.

Among various representations of the DP, we use the stick-breaking construction \citep{15sethuraman1994constructive}, which defines $P\sim \text{DP}(\alpha, P_0)$ as a weighted sum of point masses:
\begin{align}
	P(\cdot) = \sum_{k=1}^\infty \pi_k \delta_{\eta_k}(\cdot), \quad \pi_k = v_k \prod_{j<k} (1-v_j), \quad v_k \sim \text{Beta}(1, \alpha),\quad \eta_k\sim P_0,
	\label{eqn:sbp}
\end{align}
where $\pi_k$ is the weight for the $k$th degenerate distribution $\delta_{\eta_k}$ that places all probability mass at the point $\eta_k$ drawn from $P_0$ and $\text{Beta}(\cdot,\cdot)$ denotes a beta distribution. The stick-breaking construction ensures that the infinite sum of the weights adds up to 1, that is , $\sum_{k=1}^\infty \pi_k = 1$.

The stick-breaking representation reveals that a realization of the DP results in a discrete distribution, which is advantageous for serving as a prior distribution for the weights of unknown mixture components. By applying the stick-breaking construction in \eqref{eqn:sbp}, the Dirichlet process Gaussian mixture (DPGM) model can be formally defined as follows:
\begin{align}
	\begin{split}
		\mathbf{x}_i \mid z_i, \boldsymbol\mu_{z_i},\boldsymbol\Sigma_{z_i} &\sim \text N_d(\boldsymbol\mu_{z_i},\boldsymbol\Sigma_{z_i}), \quad i=1,\dots,n,\\
		z_i \mid \{v_k\}_{k=1}^\infty &\sim   \text{Discrete}(\{\pi_k\}_{k=1}^\infty), \quad i=1,\dots,n,\\
		v_k &\sim \text{Beta}(1, \alpha), \quad k=1,2,\dots ,
	\end{split}
	\label{eqn:stb}
\end{align}
where ${\text N}_d(\boldsymbol\mu,\boldsymbol\Sigma)$ denotes the $d$-dimensional Gaussian distribution with mean $\boldsymbol \mu$ and covariance matrix $\boldsymbol\Sigma$,
$\{\boldsymbol\mu_k,\boldsymbol\Sigma_k\}_{k=1}^\infty$ are sets of mean and covariance parameters of the Gaussian mixture components, $\{z_i\}_{i=1}^n$ are the mixture memberships, and $  \text{Discrete}(\{\pi_k\}_{k=1}^\infty)$ denotes a discrete probability distribution with $\Pr\{z_i = k\}=\pi_k$, $k=1,2,\dots$.

Unlike the finite GMM, the DPGM model allows for an infinite number of mixture components, eliminating the need to specify the exact number of components needed for a reasonable likelihood evaluation in outlier detection. However, this does not imply that infinitely many mixture components are used for a finite sample. Instead, the actual number of mixture components utilized is determined by the data.
For practical implementation of DPGM, setting a conservative upper bound $K$ on the maximum number of mixture components is necessary. This upper bound $K$ should be sufficiently large to accommodate the data, ensuring that its selection does not adversely affect the model. A common practice is to set $K=30$ as the truncation threshold \citep{gelman2013bayesian}.

\subsubsection{Variational inference for Dirichlet process mixtures}

Estimation of DPGM traditionally relies on MCMC methods \citep{19neal2000markov}. Despite their extensive use in complex tasks, MCMC methods often face practical challenges due to significant computational inefficiencies. As an alternative, variational inference provides a faster and more scalable algorithm for DPGM \citep{17blei2006variational}. In this study, we use variational inference owing to its computational advantages, which are particularly relevant for real-world outlier detection tasks involving large-scale datasets with high dimensionality.

Variational inference approximates the target posterior distribution with a more manageable distribution known as the variational distribution $Q$, achieved through deterministic optimization procedures \citep{jordan1999introduction}. Given instances $\mathbf X = [\mathbf{x}_1, \dots, \mathbf{x}_n]^T$, this approximation is obtained by minimizing the KL divergence between the posterior distribution $P(\cdot\mid \mathbf X)$ and the variational distribution $Q(\cdot)$:
\begin{align}
	KL(Q(\cdot),P(\cdot\mid\mathbf X)) = \int  \log q(\boldsymbol\theta)dQ(\boldsymbol\theta) -  \int  \log [p(\mathbf X \mid\boldsymbol\theta)p(\boldsymbol\theta)] dQ(\boldsymbol\theta) + \log p(\mathbf X ),
	\label{eqn:kld}
\end{align}
where $q$ is the density of $Q$, $\boldsymbol\theta$ is the set of parameters of interest, which in the case of our DPGM model are $(\{z_i\}_{i=1}^n,\{v_k,\boldsymbol\mu_k,\boldsymbol\Sigma_k\}_{k=1}^K)$, $p(\boldsymbol\theta)$ is a prior density, $p(\mathbf X \mid\boldsymbol\theta)$ is the likelihood, and $p(\mathbf X )$ is the marginal likelihood of the instances $\mathbf X$. Minimizing \eqref{eqn:kld} is equivalent to maximizing the lower bound for the marginal log-likelihood, given by
\begin{align}
	\begin{split}
		\log p(\mathbf X) 
		&\geq    \int  \log [p(\mathbf X \mid\boldsymbol\theta)p(\boldsymbol\theta)] dQ(\boldsymbol\theta) - \int  \log q(\boldsymbol\theta)dQ(\boldsymbol\theta)\\
		&= \int \log\frac{p(\mathbf X \mid\boldsymbol\theta)p(\boldsymbol\theta)}{q(\boldsymbol\theta)} dQ(\boldsymbol\theta).
	\end{split}
\end{align}
The rightmost side is commonly referred to as the \textit{evidence lower bound} (ELBO) in the literature. 
Assuming further independence of the parameters, the procedure is referred to as \textit{mean-field} variational inference \citep{jordan1999introduction}.
In this case, the variational posterior distribution is decomposed as $Q(\boldsymbol\theta)=\prod_{j=1}^b Q_j(\boldsymbol\theta_j)$, where $\boldsymbol\theta_j$ are subvectors that partition $\boldsymbol\theta$. For our DPGM model, the mean-field variational posterior takes the form  $Q(\boldsymbol\theta)=\prod_{i=1}^n Q_i^1(z_i)\prod_{k=1}^{K-1}Q_k^2(v_k)\prod_{k=1}^K Q_k^3(\boldsymbol\mu_k,\boldsymbol\Sigma_k) $.

Given the modeling assumption in \eqref{eqn:stb}, a natural prior distribution for $(\boldsymbol\mu_k,\boldsymbol\Sigma_k)$ is a normal-inverse-Wishart distribution, $(\boldsymbol\mu_k,\boldsymbol\Sigma_k)\sim\text{NIW}(\boldsymbol\xi, b , \nu, \boldsymbol\Psi)$, which implies $\boldsymbol\mu_k \mid \boldsymbol\Sigma_k \sim  \text{N}_d(\boldsymbol\xi, b^{-1} \boldsymbol\Sigma_k)$ and $\boldsymbol\Sigma_k^{-1}\sim \text{Wishart}_d(\nu,\boldsymbol\Psi^{-1})$, $k=1,\dots,K$, where $\text{Wishart}_d(\nu,\boldsymbol\Psi^{-1})$ denotes a Wishart distribution with $\nu> d-1$ degrees of freedom and a $d\times d$ positive definite scale matrix $\boldsymbol\Psi$. Selecting appropriate values for $\boldsymbol\xi$, $b$, $\nu$, $\boldsymbol\Psi$ is imperative. While setting $b=1$ and $\nu=d$, a common practice is to assign $\boldsymbol\xi$ and $\boldsymbol\Psi$ the empirical mean and covariance matrix of $\mathbf{x}_1, \dots, \mathbf{x}_n$, respectively \citep{gorur2010dirichlet}, although not purely Bayesian owing to the dependency of the prior on the data.
The induced variational posterior is obtained by direct calculations \citep{2bishop2006pattern,17blei2006variational},
\begin{align}
	\begin{split}
		z_i &\sim Q_i^1 = \text{Discrete}(\tilde\pi_{i1},\dots,\tilde\pi_{iK}), \quad i=1,\dots,n,\\
		v_k&\sim Q_k^2 = \text{Beta}(\tilde\gamma_{k1},\tilde\gamma_{k2}),\quad k=1,\dots,K-1,\\
		(\boldsymbol\mu_k , \boldsymbol\Sigma_k) &\sim Q_k^3 = \text{NIW}(\tilde{\boldsymbol\xi}_k,\tilde b_k , \tilde\nu_k, \tilde{\boldsymbol\Psi}_k),\quad k=1,\dots,K,
	\end{split}
	\label{eqn:varipos}
\end{align}
with the parameters optimized by the coordinate ascent algorithm such that
\begin{align}
	\begin{split}
		\tilde\pi_{ik} &\propto \exp\Bigg\{\psi(\tilde\gamma_{k1})+\sum_{j=1}^{K-1}\psi(\tilde\gamma_{j2})-\sum_{j=1}^K\psi(\tilde\gamma_{j1}+\tilde\gamma_{j2})+\frac{1}{2}\log \lvert\tilde{\boldsymbol\Psi}_k^{-1}\rvert\\
		&\qquad\qquad +\frac{1}{2}\sum_{j=1}^d\psi\bigg(\frac{\tilde \nu_k+1-j}{2}\bigg)-\frac{\tilde \nu_k}{2}(\mathbf x_i-\tilde{\boldsymbol\xi}_k)^T\tilde{\boldsymbol\Psi}_k^{-1}(\mathbf x_i-\tilde{\boldsymbol\xi}_k)-\frac{d}{2\tilde b_k}\Bigg\},\\
		\tilde\gamma_{k1} &= 1+\tilde n_k,\quad\tilde\gamma_{k2} =\alpha+\sum_{j=k+1}^K\tilde n_j,\\
		\tilde{\boldsymbol\xi}_k &= \frac{b\boldsymbol\xi+\tilde n_k\tilde{\mathbf w}_k}{b+\tilde n_k}, \quad
		\tilde b_k = b+\tilde n_k, \quad		\tilde \nu_k = \nu+\tilde n_k,\\
		\tilde{\boldsymbol\Psi}_k &= \boldsymbol\Psi+\sum_{i=1}^{n}\tilde\pi_{ik}(\mathbf x_i-\tilde{\mathbf w}_k)(\mathbf x_i-\tilde{\mathbf w}_k)^{T}+\frac{b \tilde n_k}{b+\tilde n_k}(\tilde{\mathbf w}_k-\boldsymbol\xi)(\tilde{\mathbf w}_k-\boldsymbol\xi)^{T},
	\end{split}
	\label{eqn:ca}
\end{align}
where $\psi$ denotes the digamma function, $\tilde n_k=\sum_{i=1}^n \tilde\pi_{ik}$, and $\tilde{\mathbf w}_k=\sum_{i=1}^n \tilde\pi_{ik} \mathbf x_i/\sum_{i=1}^n \tilde\pi_{ik}$.
To ensure that $\tilde{\boldsymbol\Psi}_k$ is positive definite, the formulation requires $d \le n$ when $\mathbf \Psi$ is chosen as the empirical covariance matrix.

Although this variational posterior provides the flexibility needed to closely approximate the target posterior distribution $P(\cdot\mid \mathbf X)$, optimizing the positive definite matrix for the inverse Wishart variational posterior of $\boldsymbol \Sigma_k$ can be time-consuming, particularly in high dimensions. An alternative approach, which is easier to train but less flexible, involves simplifying the covariance structure by setting the off-diagonal elements of $\boldsymbol \Sigma_k$ to zero in its prior distribution. In this scenario, the inverse Wishart distribution simplifies to a product of independent inverse gamma distributions. Consequently, the resulting variational posterior for $\boldsymbol \Sigma_k$ also becomes a product of independent inverse gamma distributions. This is equivalent to setting the off-diagonal elements of $\tilde{\boldsymbol\Psi}_k$ in \eqref{eqn:ca} to zero, that is, $(\tilde{\boldsymbol\Psi}_k)_{jj'} = 0$ if $1\le j\ne j'\le d$ and
\begin{align}
	(\tilde{\boldsymbol\Psi}_k)_{jj} &= (\boldsymbol\Psi)_{jj}+\sum_{i=1}^{n}\tilde\pi_{ik}(x_{ij}-\tilde{w}_{kj})^2+\frac{b \tilde n_k}{b+\tilde n_k}(\tilde{w}_{kj}-\xi_j)^2, \quad j=1,\dots,d,
	\label{eqn:ca2}
\end{align}
where $x_{ij}$, $\tilde{w}_{kj}$, and $\xi_j$ are the $j$th entries of $\mathbf x_i$, $\tilde{\mathbf w}_k$, and $\boldsymbol\xi$, respectively.
Therefore, optimization for the ELBO is more easily performed by optimizing a series of univariate variational parameters, leading to a computationally efficient algorithm that scales well to large dimensions. Notably, this simplification no longer requires $d\le n$ as the resulting $\tilde{\boldsymbol\Psi}_k$ is always positive definite.

We refer to the two covariance assumptions as the \textit{full covariance} assumption and the \textit{diagonal covariance} assumption, respectively.
Variational inference for DPGM can be implemented using the \texttt{BayesianGaussianMixture} function in \texttt{scikit-learn} with both covariance assumptions.
In clustering and density estimation tasks, the diagonal covariance assumption might significantly underperform if the data $\mathbf X$ deviates from this assumption, as it does not account for correlations between features. However, in the context of outlier detection, where subspace and subsampling ensembles (discussed in Section~\ref{sec3}) are employed, we find that diagonal covariance assumption does not compromise detection accuracy. Instead, it enhances computational efficiency, leading to reduced runtime. Therefore, we adopt the diagonal covariance assumption as the default for our outlier detection method. However, for improved visualization and understanding, all figures in this paper are generated using the full covariance assumption, which provides a more detailed representation of the data.

Once the training of the variational posterior is complete, Bayesian point estimates of the parameters can be obtained using \eqref{eqn:varipos}. Among the various options, we consider the following estimates, which are combinations of the variational posterior means and modes, as implemented in \texttt{scikit-learn}:
\begin{align}
	\begin{split}
		\hat \pi_{k} & = E_Q[v_k]\prod_{j <k }E_Q[1-v_j] = \frac{\tilde\gamma_{k1}\prod_{j< k}\tilde\gamma_{j2}}{\prod_{j\le k}(\tilde\gamma_{j1}+\tilde\gamma_{j2})},\\
		\hat z_i & = \underset{k}{\operatorname{argmax}\,} Q(z_i=k) = \underset{k}{\operatorname{argmax}\,} \tilde\pi_{ik},\\
		\hat K &= \sum_{k=1}^K \mathbbm 1\!\left\{\sum_{i=1}^n \mathbbm 1\{\hat z_i=k\}>0\right\},\\
		\hat {\boldsymbol\mu}_k &= E_Q[{\boldsymbol\mu}_k] = \tilde{\boldsymbol \xi}_k,\\
		\hat {\boldsymbol\Sigma}_k &= E_Q[{\boldsymbol\Sigma}_k^{-1}]^{-1} =\tilde \nu_k^{-1} \tilde {\boldsymbol \Psi}_k,
	\end{split}
	\label{eqn:estimates}
\end{align}
where $Q$ represents the variational posterior as defined in \eqref{eqn:varipos}, and $E_Q$ denotes the corresponding expectation operator. Specifically, $\hat \pi_{k}$, $\hat {\boldsymbol\mu}_k$, and $\hat {\boldsymbol\Sigma}_k$ are obtained using the posterior means of $\{v_j\}_{j\le k}$, ${\boldsymbol\mu}_k$, and ${\boldsymbol\Sigma}_k^{-1}$, respectively. Note that to estimate ${\boldsymbol\Sigma}_k$, we use the posterior mean of its inverse, which corresponds to the mean of the Wishart variational posterior. Although using the inverse Wishart distribution directly is possible, the form in \eqref{eqn:estimates} is preferred because it strikes a balance between the posterior mean and the mode of the inverse Wishart variational posterior, thereby reducing sensitivity. Since $z_i$ is discrete, the estimate $\hat z_i$ is determined by its posterior mode. Finally, $\hat K\le K$ represents the number of mixture components to which at least one instance is assigned.

\subsection{Outlier ensemble for the proposed method}
\label{sec3}

Our proposed method for outlier detection leverages two key concepts in ensemble construction: subspace and subsampling ensembles. These approaches aim to reduce the training dataset $\mathbf{D} \in \mathbb{R}^{N \times p}$ by creating $M$ ensemble components $\mathbf{X}_m\in\mathbb R^{n_m\times d_m}$, $m = 1, \dots, M$, which are then averaged to construct an outlier detector. Specifically, subspace projection is used to reduce the number of features from $p$ to $d_m$, while subsampling decreases the number of instances from $N$ to $n_m$.

\subsubsection{Subspace ensemble}
\label{sec:subspa}

In our proposed method, the original training dataset is randomly projected onto subspaces of dimensions smaller than $p$ to generate multiple ensemble components. 
The use of random projection is justified by the Johnson-Lindenstrauss lemma, which states that orthogonal projections preserve pairwise distances with high probability \citep{johnson1984extensions}. Additionally, random projection facilitates the Gaussian mixture modeling of non-Gaussian data owing to the central limit theorem \citep{diaconis1984asymptotics}. Specifically, for each $m=1,\dots,M$, the training dataset $\mathbf D\in\mathbb R^{N\times p}$ is projected onto $d_m$-dimensional subspace as $\mathbf D\mathbf R_m\in\mathbb R^{N\times d_m}$ using a random projection matrix $\mathbf R_m\in\mathbb R^{p\times d_m}$, where $d_m\le p$. Several methods can be used to generate a random projection matrix $\mathbf R_m$. A true random projection is obtained by choosing columns as $d$-dimensional random orthogonal unit-length vectors \citep{johnson1984extensions}. However, computationally efficient alternatives exist, such as drawing entries from the standard Gaussian distribution or discrete distributions without orthogonalization \citep{blum2005random}. 
In our approach, we  generate a random projection matrix $\mathbf R_m$ by drawing each element from a uniform distribution over $(-1, 1)$ and then applying Gram-Schmidt orthogonalization to the columns.

\begin{figure}[t!]
	\centering
	\includegraphics[width = 13cm]{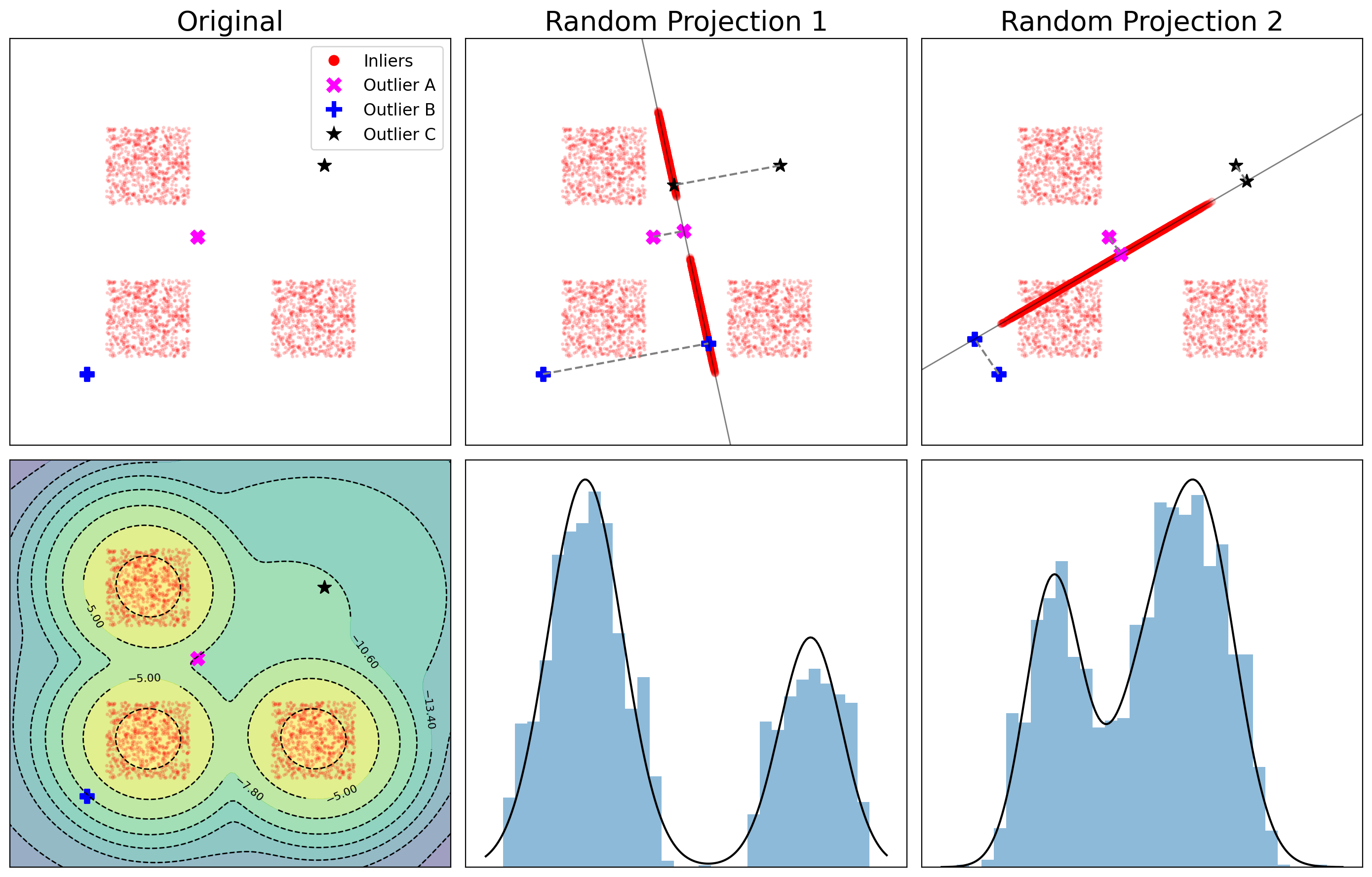}
	\caption{Random projections of two-dimensional data onto one dimension. In the upper panels, outlier A becomes evident through the first random projection (middle), while outliers B and C are captured using the second random projection axis (right).
		In the lower panels, the original two-dimensional data exhibit a significant bias when modeled with Gaussian mixtures (left). The one-dimensional projected datasets are shown to be more suitable for Gaussian mixture modeling.}
	\label{plot:fig2}
\end{figure}

To illustrate the benefits of subspace outlier ensembles for the proposed method, Figure~\ref{plot:fig2} shows a simulated two-dimensional dataset projected onto one-dimensional random rotation axes. (As clarified below, our proposed method performs dimension reduction only when $p\ge 3$; the toy example is included solely for graphical purposes to highlight the effectiveness of the subspace ensemble.) The figure demonstrates how various random projections reveal each of the three outliers. By using an ensemble of one-dimensional random projections, all three outliers can be effectively detected.
The figure also showcases the advantages of random projection within the Gaussian mixture framework. The original two-dimensional dataset comprises three main clusters arranged in squares, which deviate significantly from Gaussian distributions. Fitting the data in its original dimension introduces considerable bias, as illustrated. However, one-dimensional random projections make the data more amenable to Gaussian modeling with a few mixture components.
This example highlights the effectiveness of subspace outlier ensembles in uncovering outliers that might be hidden in a single projection or in a full-dimensional analysis.

\begin{figure}[t!]
	\centering
	\includegraphics[width = 11cm]{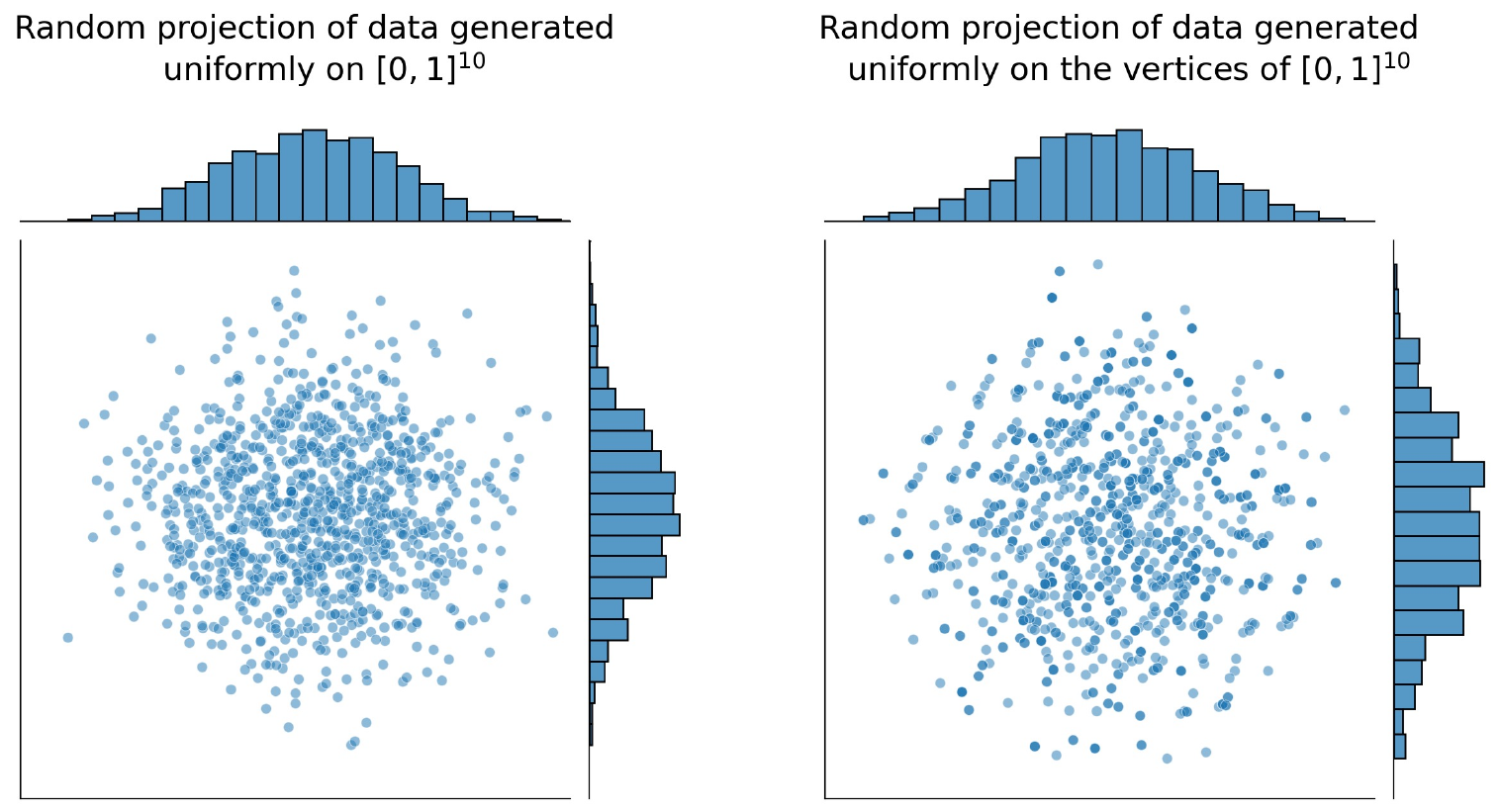}
	\caption{Two-dimensional random projections of ten-dimensional non-Gaussian data of size $1000$.  The left panel shows a random projection of data generated uniformly on the unit cube $[0,1]^{10}$, i.e., $\prod_{j=1}^{10}\text{Uniform}(0,1)$. The right panel shows a random projection of data generated uniformly on the vertices of the unit cube $[0,1]^{10}$, i.e., $\prod_{j=1}^{10}\text{Bernoulli}(1/2)$.}
	\label{plot:rp} 
\end{figure}

Figure~\ref{plot:rp} further illustrates the effectiveness of random projection combined with Gaussian mixture modeling for non-Gaussian data. The left panel illustrates a two-dimensional random projection of ten-dimensional data randomly generated within the unit cube $[0,1]^{10}$. Despite the original data's non-Gaussian nature in ten dimensions, the projected data closely resemble a two-dimensional Gaussian distribution.
In the right panel, we observe a two-dimensional random projection of ten-dimensional data randomly generated on the vertices of the unit cube $[0,1]^{10}$. Here, the original data are not only non-Gaussian but also discrete, with $2^{10}$ possible values. Similar to the first case, the projected data appear well-suited for Gaussian modeling. This suggests that using Gaussian mixture models with random projection is effective even for discrete datasets.

Determining the most appropriate subspace dimensions $d_m$ remains challenging. Insufficient dimensionality may fail to capture the data's overall characteristics, while excessive dimensionality can undermine the benefits of subspace ensembles. One common strategy is to choose the subspace dimension $d_m$ as a random integer between $\min\{p,2 + \sqrt{p}/2\}$ and $\min\{p,2 + \sqrt{p}\}$. (Accordingly, dimension reduction occurs only when $p \ge 3$.)
This strategy, based on \cite{35aggarwal2015}, is grounded in the observation that the informative dimensionality of most real-world datasets typically does not exceed $\sqrt{p}$.

\subsubsection{Subsampling ensemble}
\label{sec:subsam}

As previously discussed, a primary challenge in using mixture models for outlier detection is their high computational cost. Training these models involves assigning instances to cluster memberships, which becomes extensive when the training dataset $\mathbf D \in \mathbb{R}^{N \times p}$ includes a large number of instances $N$. The training process is time-consuming because each data point requires membership determination. For instance, in the case of DPGM with MCMC, all membership indicators must be updated in every MCMC iteration. Although we employ variational inference for DPGM to improve computational efficiency, challenges persist in the optimization procedure. Since determining cluster memberships is the most computationally intensive aspect, overall computation time scales with the number of instances. This scaling makes managing large datasets for mixture models in outlier detection challenging. In this context, subsampling--randomly drawing $n_m$ instances from the training data without replacement--proves highly effective in reducing computation time.

\begin{figure}[t!]
        \centering
	\includegraphics[width = 11cm]{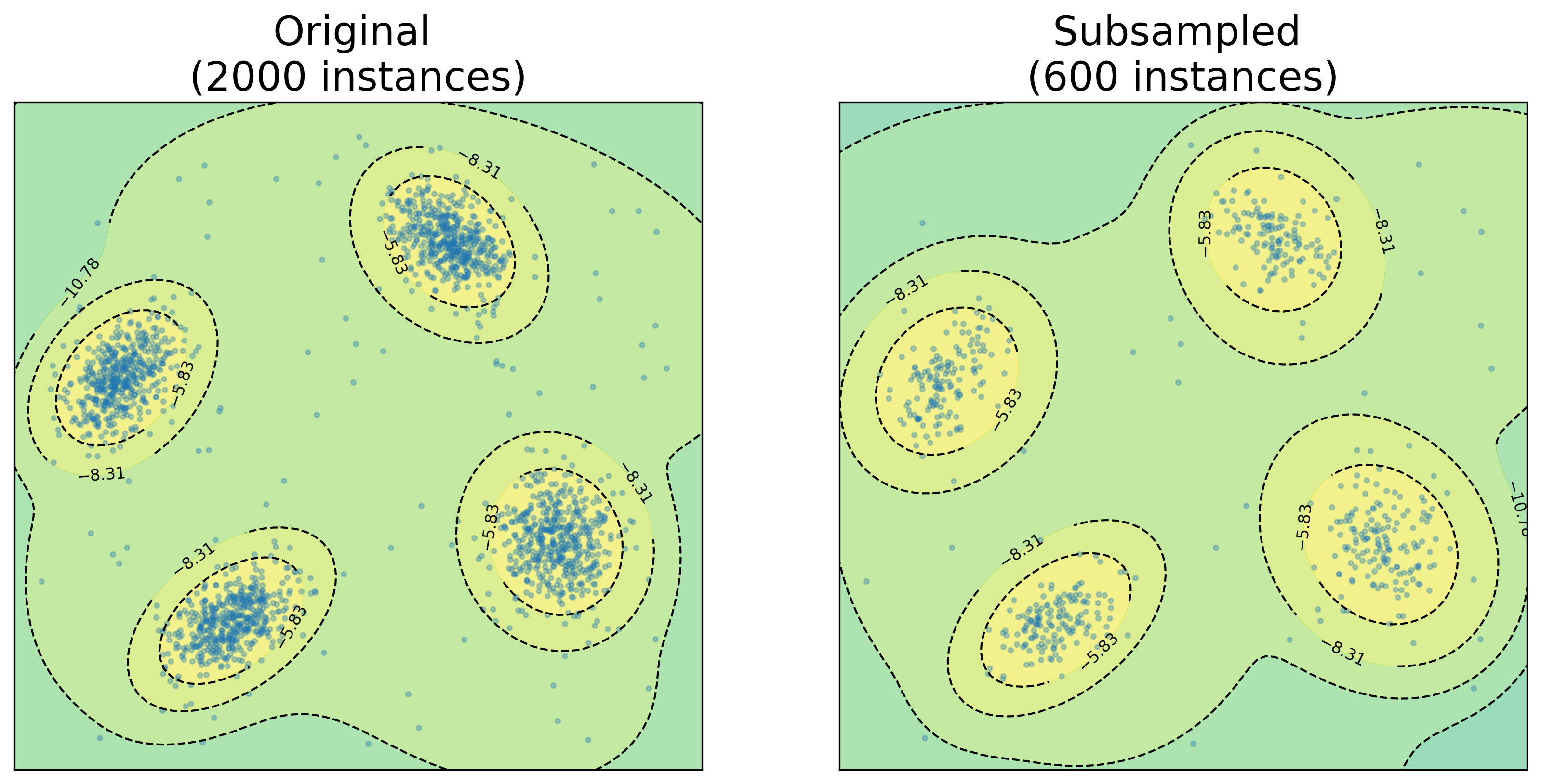}
	\caption{Log-densities estimated by DPGM for the original data (left) and the subsampled data (right). The patterns in the original data are well captured by the subsampled data.}
	\label{plot:fig4}
\end{figure}

Despite its significant computational benefits, subsampling does not compromise outlier detection accuracy when using DPGM. Figure~\ref{plot:fig4} illustrates the log-densities of both the original and subsampled data as estimated by DPGM. The original dataset consists of four main clusters with several outliers interspersed. The two estimated densities are sufficiently close, indicating 
that the subsampled dataset effectively captures the patterns of the original data. Thus, subsampling proves highly effective in reducing computation time while maintaining detection accuracy.

Similar to subspace ensembles, the optimal subsample size $n_m$ is not precisely known. A practical approach is to introduce variability in the subsample size for each ensemble component. Following \cite{35aggarwal2015}, we randomly select $n_m$ as an integer between $\min\{N,50\}$ and $\min\{N,1000\}$. Although this strategy is termed `variable subsampling' in \cite{35aggarwal2015}, we refer to it simply as `subsampling' throughout this paper. This approach ensures that subsample sizes range between 50 and 1000 when $N\ge1000$, meaning the subsample sizes are not directly proportional to the original data size. While this might seem to overlook the benefits of larger data sizes, \cite{35aggarwal2015} noted that a subsample size of 1000 is generally sufficient to model the underlying distribution of the original data. This strategy performs well even with large datasets, as an ensemble of small subsamples reduces correlation between components, thereby enhancing the benefits of the ensemble method. Our experience with DP mixture modeling for outlier detection supports this conclusion. Notably, if the full covariance assumption is used, the requirement $d_m\le n_m$ is strictly enforced with the empirical covariance matrix for $\mathbf \Psi$. This is another reason why the diagonal covariance assumption is preferred.

\section{Proposed algorithm}
\label{sec4}

\begin{figure}[!tb]
        \centering
	\includegraphics[width = 16cm]{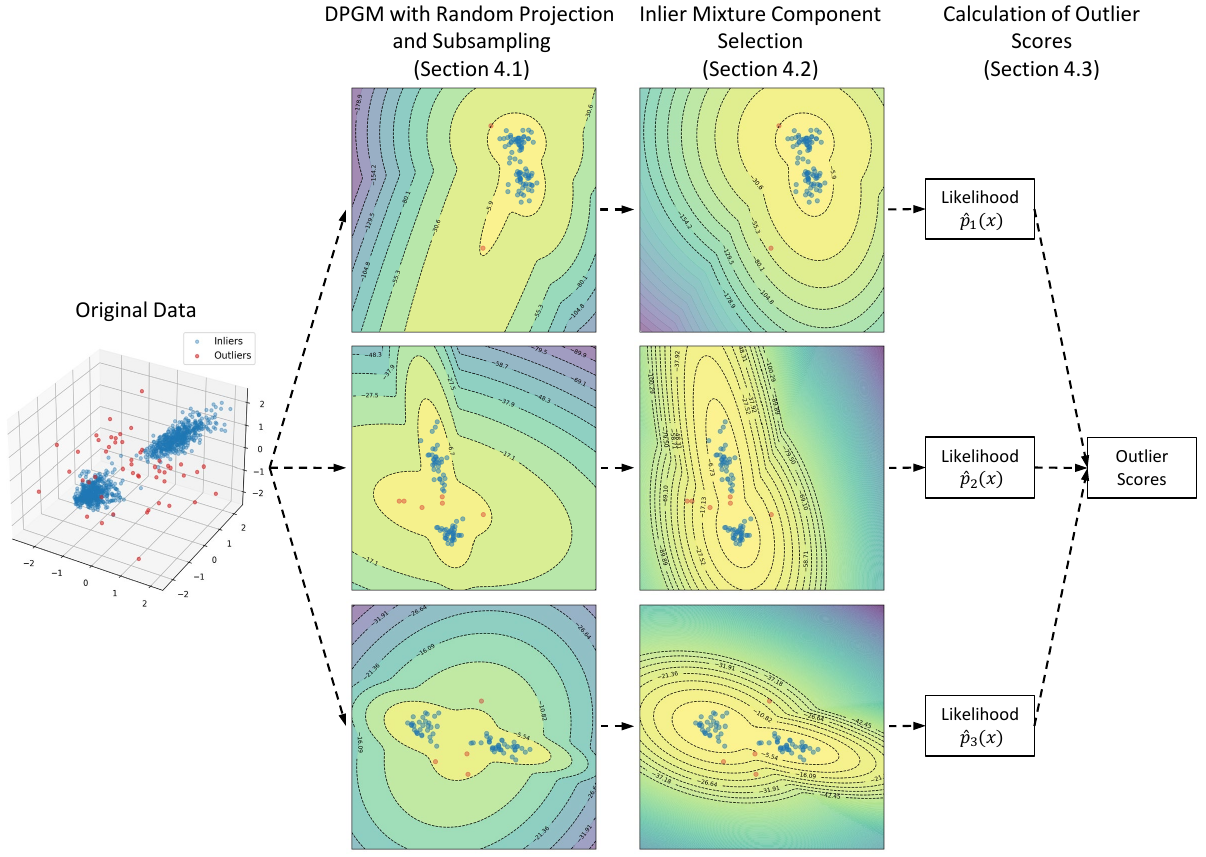}
	\caption{Graphical illustration of OEDPM. The three-dimensional original dataset is fitted using variational DPGM with random projection and subsampling (see Section~\ref{VR}). Mixture components with small weights are excluded to minimize the influence of outliers when constructing outlier detectors (see Section~\ref{inlier selection}). After pruning the mixture components, outlier scores are calculated based on the likelihood (see Section~\ref{outlier score}). }
	\label{plot:method}
\end{figure}

This section details the proposed method, referred to as the outlier ensemble of Dirichlet process mixtures (OEDPM), highlighting its unique properties and considerations for outlier detection.
The OEDPM algorithm operates through a three-step process for each ensemble component. First, it estimates the density function for reduced data using DPGM coupled with mean-field variational inference. Second, it reduces the influence of outliers in density estimation by discarding mixture components with insignificant posterior mixture weights. Third, it calculates the likelihood values of individual instances by evaluating the estimated density function at each respective data point. Outlier scores for individual instances are then obtained by aggregating likelihood values across all ensemble components. These three procedural steps are elaborated in Sections~\ref{VR}, \ref{inlier selection}, and \ref{outlier score}, respectively. Figure~\ref{plot:method} provides an illustrative example demonstrating the procedural sequence. The algorithm of OEDPM is outlined in Algorithm~\ref{alg:oedpm}.

\begin{algorithm}[p!]
	\label{alg:oedpm}
	\caption{Outlier Ensemble of Dirichlet Process Mixtures (OEDPM)}
	{\footnotesize
	\KwInput{Training dataset $\mathbf D\in\mathbb R^{N\times p}$, test dataset $\mathbf{D}^{\text{test}}\in\mathbb{R}^{N^{\text{test}}\times p}$, and contamination parameter $\phi\in(0,1)$.}
	\KwOutput{Outlier scores $O_1,\dots,O_{N_\text{test}}$ and  memberships $I_1,\dots,I_{N_\text{test}}$ of test dataset $\mathbf{D}^{\text{test}}\in\mathbb{R}^{N^{\text{test}}\times p}$.}
	
	\For{$m=1,\dots,M$} {
		\tcp{Generating ensemble components}
		
		Draw a random integer $d_m$ between $\min\{p,2 + \sqrt{p}/2\}$ and $\min\{p,2 + \sqrt{p}\}$;
		
		Generate a random projection matrix $\mathbf R_m\in\mathbb R^{p \times d_m}$ by drawing each element from $\text{Uniform}(-1, 1)$ and then applying the Gram-Schmidt process to the columns;
		
		Draw a random integer $n_m$ between $\min\{N,50\}$ and $\min\{N,1000\}$;
		
		Generate $\mathbf D_m\in\mathbb R^{n_m\times p}$ by drawing $n_m$ rows without replacement from $\mathbf D$;
		
		Generate $\mathbf X_m= [\mathbf x_{m1},\dots,\mathbf x_{m n_m}]^T = \mathbf D_m \mathbf R_m \in\mathbb R^{n_m \times d_m}$;

		\tcp{Mean-field variational inference of DPGM (easily performed by \texttt{scikit-learn})}
		
		Set hyperparameters $\alpha=1$, $b_m=1$, $\nu_m=d_m$, $\boldsymbol \xi_m = \bar{\mathbf x}_m =n_m^{-1} \sum_{i=1}^{n_m} \mathbf x_{mi}$, and $\boldsymbol \Psi_m =n_m^{-1} \sum_{i=1}^{n_m} (\mathbf x_{mi}-\bar{\mathbf x}_m)(\mathbf x_{mi}-\bar{\mathbf x}_m)^T$;
		
		Initialize variational parameters $\{\{\tilde\pi_{mik}\}_{i=1}^{n_m},\tilde\gamma_{mk1},\tilde\gamma_{mk2},\tilde{\boldsymbol\xi}_{mk},\tilde b_{mk},\tilde \nu_{mk},\tilde{\boldsymbol\Psi}_{mk}\}_{k=1}^K$;
		
		\If{diagonal covariance assumption}{
				Set all off-diagonal elements of $\tilde{\boldsymbol\Psi}_{mk}$ to zero;
		}
		
		\While{not converged}
		{
		\For{$i=1,\dots,n_m$} {
		\For{$k=1,\dots,K$} {
			Update $\tilde\pi_{mik} \leftarrow \exp\Big\{\psi(\tilde\gamma_{mk1})+\sum_{j<k}\psi(\tilde\gamma_{mj2})-\sum_{j\le k}\psi(\tilde\gamma_{mj1}+\tilde\gamma_{mj2})+\frac{1}{2}\log \lvert\tilde{\boldsymbol\Psi}_{mk}^{-1}\rvert +\frac{1}{2}\sum_{j=1}^{d_m}\psi\big(\frac{\tilde \nu_{mk}+1-j}{2}\big)-\frac{\tilde \nu_{mk}}{2}(\mathbf x_{mi}-\tilde{\boldsymbol\xi}_{mk})^T\tilde{\boldsymbol\Psi}_{mk}^{-1}(\mathbf x_{mi}-\tilde{\boldsymbol\xi}_{mk})-{d_m}/({2\tilde b_{mk}})\Big\}$;
		}
		\For{$k=1,\dots,K$} {
			Normalize $\tilde\pi_{mik}\leftarrow \tilde\pi_{mik}/\sum_{k=1}^K\tilde\pi_{mik}$;
		}
	}
	\For{$k=1,\dots,K$} {		
		Set $\tilde n_{mk}=\sum_{i=1}^{n_m} \tilde\pi_{mik}$ and $\tilde{\mathbf w}_{mk}=\sum_{i=1}^{n_m} \tilde\pi_{mik} \mathbf x_{mi}/\sum_{i=1}^{n_m} \tilde\pi_{mik}$;
		
			Update $\tilde\gamma_{mk1} \leftarrow 1+\tilde n_{mk}$ and $\tilde\gamma_{mk2} \leftarrow \alpha+\sum_{j>k}\tilde n_{mj}$;
			
			Update $\tilde{\boldsymbol\xi}_{mk} \leftarrow ({b_m\boldsymbol\xi_m+\tilde n_{mk}\tilde{\mathbf w}_{mk}})/({b_m+\tilde n_{mk}})$, $\tilde b_{mk} \leftarrow b_m+\tilde n_{mk}$, and $\tilde \nu_{mk} \leftarrow \nu_m+\tilde n_{mk}$;
			
		\uIf{diagonal covariance assumption}{
		
		\For{$j=1,\dots,d_m$} {
			Update $(\tilde{\boldsymbol\Psi}_{mk})_{jj}\leftarrow (\boldsymbol\Psi_m)_{jj}+\sum_{i=1}^{n_m}\tilde\pi_{mik}(x_{mij}-\bar{x}_{mkj})^2+\frac{b_m \tilde n_{mk}}{b_m+\tilde n_{mk}}(\tilde{w}_{mkj}-\xi_{mj})^2$;
		}
	}
					\ElseIf{full covariance assumption}
	{
		Update $\tilde{\boldsymbol\Psi}_{mk}\leftarrow \boldsymbol\Psi_m+\sum_{i=1}^{n_m}\tilde\pi_{mik}(\mathbf x_{mi}-\tilde{\mathbf w}_{mk})(\mathbf x_{mi}-\tilde{\mathbf w}_{mk})^{T}+\frac{b_m \tilde n_{mk}}{b_m+\tilde n_{mk}}(\tilde{\mathbf w}_{mk}-\boldsymbol\xi_m)(\tilde{\mathbf w}_{mk}-\boldsymbol\xi_m)^{T}$;
	}

		}
		}
	
	\tcp{Calculating outlier scores and outlier memberships using the likelihood}
	\For{$k=1,\dots,K$} {
		Set $\hat \pi_{mk} = {(\tilde\gamma_{mk1}\prod_{j< k}\tilde\gamma_{mj2})}/{\prod_{j\le k}(\tilde\gamma_{mj1}+\tilde\gamma_{mj2})}$, $\hat {\boldsymbol\mu}_{mk} = \tilde{\boldsymbol \xi}_{mk}$, and $\hat {\boldsymbol\Sigma}_{mk} =\tilde \nu_{mk}^{-1} \tilde {\boldsymbol \Psi}_{mk}$;
	}
Set $\hat K_m = \sum_{k=1}^K \mathbbm 1\!\left\{\sum_{i=1}^{n_m} \mathbbm 1\!\left\{ \operatorname{argmax}_j \tilde\pi_{mij}=k\right\}>0\right\}$;

	Set $\mathcal K_m^\ast=\{1\le k \le K : \hat\pi_{mk} \ge 1/\hat K_m\text{ or }\hat\pi_{mk} = \max_j\hat\pi_{mj}\}$;

Calculate the density $\hat p_m(\cdot) = \sum_{k\in \mathcal K_m^\ast} \hat\pi_{mk} {\varphi}_{d_m}(\cdot \, ;\hat{\boldsymbol \mu}_{mk}, \hat{\boldsymbol\Sigma}_{mk})/\sum_{k\in\mathcal K_m^\ast} \hat\pi_{mk}$;

\uIf{IQR method = True}{
Calculate $T_m=\text{Q1}_m-1.5\times \text{IQR}_m$, where $\text{Q1}_m$ and $\text{IQR}_m$ are the first quartile and the interquartile range of $\log \hat p_m(\mathbf x_{m1}),\dots,\log \hat p_m(\mathbf x_{m n_m})$;
}
\ElseIf{IQR method = False}{
Calculate $T_m$ as the $100\phi\%$ quantile of $\log \hat p(\mathbf x_{m1}),\dots,\log \hat p(\mathbf x_{m n_m})$;
}

Generate $\mathbf X_m^{\text{test}} =[\mathbf x_{m1}^{\text{test}},\dots,\mathbf x_{m n_m}^{\text{test}}]^T= \mathbf{D}^{\text{test}} \mathbf R_m\in\mathbb R^{N^{\text{test}}\times d_m}$;
	}
\For{$i=1,\dots,N^{\text{test}}$} {		
	Calculate $O_i = M^{-1}\sum_{m=1}^M \mathbbm 1\{\log \hat p_m(\mathbf x^{\text{test}}_{m i})< T_m\}$ and $I_i = \mathbbm 1\{O_i>1/2\}$;
	}

} 
\end{algorithm}

\subsection{DPGM with subspace and subsampling ensembles}
\label{VR}

Based on the discussion in Section~\ref{sec3}, our proposed OEDPM leverages the advantages of outlier ensembles by incorporating random projection and subsampling techniques. The resulting reduced dataset, after being subsampled and projected, is then trained using DPGM with variational inference. By repeating this process, we generate $M$ ensemble components. These components are subsequently combined to assess whether each instance in the full dataset is an outlier. The procedure is summarized as follows.

\begin{enumerate}
	\item For $d_m$, chosen as a random integer between $\min\{p,2 + \sqrt{p}/2\}$ and $\min\{p,2 + \sqrt{p}\}$, generate a random projection matrix $\mathbf R_m\in\mathbb R^{p \times d_m}$, where each element is sampled randomly from $\text{Uniform}(-1, 1)$ and the columns are orthogonalized through the Gram-Schmidt process.
	
	\item For $n_m$, chosen as a random integer between $\min\{N,50\}$ and $\min\{N,1000\}$, randomly draw $n_m$ instances without replacement from $\mathbf D\in\mathbb R^{N\times p}$ to form a reduced dataset $\mathbf D_m\in\mathbb R^{n_m\times p}$.
	
	\item Produce $\mathbf X_m = \mathbf D_m \mathbf R_m\in\mathbb R^{n_m \times d_m}$ to generate a reduced dataset projected onto a random subspace. 
	\item Fit DPGM to $\mathbf X_m$ to obtain the variational posterior distribution in \eqref{eqn:varipos} using the updating rules in \eqref{eqn:ca} and \eqref{eqn:ca2} with additional subscripts $m$ (see Algorithm~\ref{alg:oedpm}).
\end{enumerate}

Our method emphasizes computational feasibility while reducing the risk of overfitting that can occur with a single original dataset. This ensemble approach reduces variance by leveraging the diversity of base detectors \citep{35aggarwal2015}. Additionally, the reduced dimensionality and size of the training data result in significant computational savings, effectively addressing concerns associated with probabilistic mixture models.

\subsection{Inlier mixture component selection}
\label{inlier selection}

The fundamental principle of outlier detection involves analyzing normal patterns within a dataset. The success of an outlier detection method depends on how well it models the inlier instances to identify points that deviate from these normal instances. 
In unsupervised outlier detection, we work with contaminated datasets where normal instances are mixed with noise and potential outliers \citep{garcia2011tclust,punzo2016parsimonious}. 
An effective algorithm should filter out outliers during training. While the DPGM has the advantage of automatically determining the optimal number of mixture components, it can also be problematic as it may overfit to anomaly instances. To address this issue, we prune irrelevant mixture components based on the posterior information of the mixture weights in DPGM.

Outliers are typically isolated instances that deviate significantly from the majority of data points. Therefore, a natural assumption is that outliers will not conform to any existing cluster memberships, resulting in less stable clusters with fewer instances. For each ensemble component $m=1,\dots,M$, let $\hat K_m\le K$ be the number of mixture components and $\{\hat\pi_{m1},\dots,\hat\pi_{mK}\}$ be the mixture weights estimated by DPGM as in \eqref{eqn:estimates} with additional subscripts $m$ (see Algorithm~\ref{alg:oedpm}). We discard mixture components with insignificant posterior weights to redefine the model and enhance its ability to detect outliers.
If no mixture component has $\hat\pi_{mk} \ge 1/\hat K_m$, only the component with the largest $\hat\pi_{mk}$ is retained. This results in a more robust mixture distribution that comprise only inlier Gaussian components, effectively filtering out outliers from the inlier set. This pruning process is applied to all ensemble components for $m=1,\dots,M$. Our experience indicates that selecting inlier mixture components is crucial for achieving reasonable performance in OEDPM. The advantages of this pruning step are clearly illustrated in Figure~\ref{plot:method}.

\subsection{Calculation of outlier scores}
\label{outlier score}

DPGM is widely recognized as a probabilistic clustering method, with each identified cluster potentially serving as a criterion for outlier detection \citep{56shotwell2011bayesian}. Specifically, from a clustering perspective, instances that do not align with the predominant clusters can be considered outliers. However, relying solely on cluster memberships to identify outliers may not be ideal, as DPGM provides a global characteristic of the entire dataset through the likelihood, which differs from proximity-based clustering algorithms. Thus, computing outlier scores based on the likelihood, rather than relying solely on cluster memberships, is more appropriate.

Given the trained $m$th ensemble component, the likelihood of an instance $\mathbf x\in\mathbb R^{d_m}$ is expressed as 
\begin{align}
	\hat p_m(\mathbf x) = \sum_{k\in \mathcal K_m^\ast} \hat\pi_{mk}^\ast {\varphi}_{d_m}(\mathbf x ;\hat{\boldsymbol \mu}_{mk}, \hat{\boldsymbol\Sigma}_{mk}),
\end{align}
where $\mathcal K_m^\ast=\{1\le k \le K : \hat\pi_{mk} \ge 1/\hat K_m\text{ or }\hat\pi_{mk} = \max_j\hat\pi_{mj}\}$ is the index set for mixture components after pruning, as described in Section~\ref{inlier selection}, $\hat\pi_{mk}^\ast$ is the weight renormalized from $\hat\pi_{mk}$ such that $\sum_{k\in\mathcal K_m^\ast} \hat\pi_{mk}^\ast = 1$ with the pruned mixture components, 
and $\hat{\boldsymbol \mu}_{mk}$ and $\hat{\boldsymbol\Sigma}_{mk}$ are the parameters estimated by DPGM as in \eqref{eqn:estimates} with additional subscript $m$ (see Algorithm~\ref{alg:oedpm}).
A relatively small likelihood value of an instance suggests it could potentially be an outlier. To define the outlier score, we need to consider a threshold that assigns a binary score to a test instance within a reduced subspace.
We examine the following two methods of obtaining this threshold.

\begin{itemize}
	\item A contamination parameter $\phi\in(0,1)$ can be used to construct the outlier scores. This method is particularly useful if the proportion of outliers in the dataset is roughly known. For a given $\phi\in(0,1)$, we define the cut-off threshold $T_m$ as the $100\phi\%$ quantile of $\log \hat p_m(\mathbf x_{m1}),\dots,\log \hat p_m(\mathbf x_{m n_m})$, where $\mathbf x_{mi}$ is the $i$th row (instance) of $\mathbf X_m$.
	\item Establishing a threshold without a contamination parameter is also feasible. This approach may be beneficial when the proportion of outliers is entirely unknown. Specifically, we define the cut-off threshold $T_m$ as $T_m=\text{Q1}_m-1.5\times \text{IQR}_m$ using the rule of thumb, where $\text{Q1}_m$ and $\text{IQR}_m$ are the first quartile and the interquartile range (IQR) of $\log \hat p_m(\mathbf x_{m1}),\dots,\log \hat p_m(\mathbf x_{m n_m})$, respectively. 
\end{itemize}

The IQR method is appealing because it does not require a user-specified contamination parameter. However, as observed in Section~\ref{sec:sens} with benchmark datasets, while the IQR method is often satisfactory, it can occasionally underperform compared to methods using a manually determined contamination parameter, such as $\phi=0.1$.

Let $\mathbf{D}^{\text{test}}\in\mathbb{R}^{N^{\text{test}}\times p}$ represent a test dataset. This dataset can coincide with the original dataset $\mathbf D$ if the interest is in identifying outliers within the provided data, or it can consist of entirely new data collected separately.
Using the random projection matrices $\mathbf R_m\in\mathbb R^{p\times d_m}$, $m=1,\dots,M$, used for training, the test instances projected onto the subspaces are expressed as
$\mathbf X_m^{\text{test}}= \mathbf{D}^{\text{test}} \mathbf R_m\in\mathbb R^{N^{\text{test}}\times d_m}$, $m=1,\dots,M$.
With the threshold $T_m$ defined by either method, we calculate the outlier score of each test instance using binary thresholding based on their likelihood values:
\begin{align}
	O_i = \frac{1}{M}\sum_{m=1}^M \mathbbm 1\{\log \hat p_m(\mathbf x^{\text{test}}_{m i})< T_m\},\quad i=1,\dots,N^{\text{test}},
	\label{eqn:score}
\end{align}
where $\mathbf x^{\text{test}}_{m i}$ is the $i$th row of $\mathbf X_m^{\text{test}}$. Using the outlier scores, we calculate outlier membership indicators $I_i = \mathbbm 1\{O_i>1/2\}$. Therefore, a voting classifier categorizes $\mathbf x^{\text{test}}_{m i}$ as an outlier if $O_i>1/2$ and as an inlier otherwise. 
Even when a specific contamination parameter $\phi\in(0,1)$ is used to determine the thresholds $T_m$, the resulting estimate of the outlier proportion is not necessarily identical to $\phi$ because the identified outliers are determined by the rule $O_i>1/2$ averaged over all ensemble components. This introduces some degree of robustness against the specification of $\phi$.

Additionally, instead of the binary thresholding $\mathbbm 1\{\log\hat p_m(\mathbf x^{\text{test}}_{m i})< T_m\}$ used in \eqref{eqn:score}, one might also define the outlier score directly using the magnitude of the likelihood values, for example, $T_m-\log\hat p_m(\mathbf x^{\text{test}}_{m i})$. However, our observations reveal that using binary thresholding significantly improves the stability and robustness of our outlier detection task.

\section{Numerical results}
\label{sec:data}

\subsection{Sensitivity analysis}
\label{sec:sens}

We evaluate the performance of OEDPM using the benchmark datasets available in the ODDS library (\url{https://odds.cs.stonybrook.edu}). We include 27 multi-dimensional point datasets with outlier labels, as four of the 31 datasets listed are incomplete. The datasets are categorized into continuous or discrete types based on the nature of the instance values, with some datasets presenting a mix of both. Details  of the benchmark datasets are summarized in Table~\ref{tab:data}.

\begin{table}[t!]
	\centering
	\resizebox{0.8\width}{!}{%
		\begin{tabular}{ lrrrr }
			\toprule
			Dataset & Size ($N$) &  Dimension ($p$) & \# of outliers  ($\%$) & Type \\
			\midrule
			Smtp (KDDCUP99) & 95156 & 3 & 30 $(0.03\%)$ & Continuous \\
			Http (KDDCUP99) & 567479 & 3 & 2211 $(0.39\%)$ & Continuous \\
			ForestCover & 286048 & 10 & 2747 $(0.96\%)$ & Continuous \\
			Satimage & 5803 & 36 & 71 $(1.22\%)$ & Continuous \\
			Speech & 3686 & 400 & 61 $(1.65\%)$ & Continuous \\
			Pendigits & 	6870 &	16	& 156 $(2.27\%)$ & Continuous \\
			Mammography & 11183 & 6 & 260 $(2.32\%)$ & Continuous \\
			Thyroid & 3772 & 6 & 93 $(2.47\%)$ & Continuous \\
			Optdigits & 5216 &	64	& 150 $(2.88\%)$ & Continuous \\
			Musk & 3062 & 166 & 97 $(3.17\%)$ & Mixed \\
			Vowels & 1456 &	12	& 50 $(3.43\%)$ & Continuous\\
			Lympho & 148 & 18 & 6 $(4.05\%)$ & Discrete \\
			Glass & 214 & 9	& 9 $(4.21\%)$ & Continuous \\
			WBC & 378 & 30 & 21 $(5.56\%)$ & Continuous \\
			Letter Recognition & 1600 &	32	& 100 $(6.25\%)$ & Continuous \\
			Shuttle & 49097 & 9 & 3511 $(7.15\%)$ & Mixed \\
			Annthyroid & 7200 & 6 & 534 $(7.42\%)$ & Continuous \\
			Wine & 129 & 13 & 10 $(7.75\%)$ & Continuous \\
			Mnist & 7603 & 100 & 700 $(9.21\%)$ & Continuous \\
			Cardio & 1831 & 21 & 176 $(9.61\%)$ & Continuous \\
			Vertebral & 240 & 6 & 30 $(12.50\%)$ & Continuous \\
			Arrhythmia & 452 & 274 & 66 $(14.60\%)$ & Mixed \\
			Heart & 267 & 44 & 55 $(20.60\%)$ & Continuous \\
			Satellite & 6435 & 36 & 2036 $(31.64\%)$ & Continuous \\
			Pima & 768 & 8 & 268 $(34.90\%)$ & Mixed \\
			BreastW & 683 & 9 & 239 $(34.99\%)$ & Discrete \\
			Ionosphere & 351 & 33 & 126 $(35.90\%)$ & Mixed \\
			\bottomrule
		\end{tabular}
	}
	\caption{Details of the 27 benchmark datasets, sorted in ascending order based on the proportion of outliers. The dataset type is labeled as `Continuous' when all features are continuous, `Discrete' when all features are discrete, and `Mixed' otherwise.}
	\label{tab:data}
\end{table}

To examine the sensitivity of the contamination parameter $\phi$, we apply OEDPM to each benchmark dataset with $M=100$ and various values of $\phi$. For this numerical analysis, the test dataset is the same as the training dataset, and all datasets are standardized before analysis.
For each specific value of $\phi$, outlier scores $O_i$ are computed for every instance. Instances are classified as outliers if their corresponding $O_i$ values are greater than $1/2$. We then calculate the F1-scores based on these outlier detection results. Additionally, we compute the F1-scores using the IQR-based strategy outlined in Section~\ref{outlier score}, which does not require specifying a value for $\phi$.

\begin{figure}[t!]
	\centering
	\includegraphics[width = 16cm]{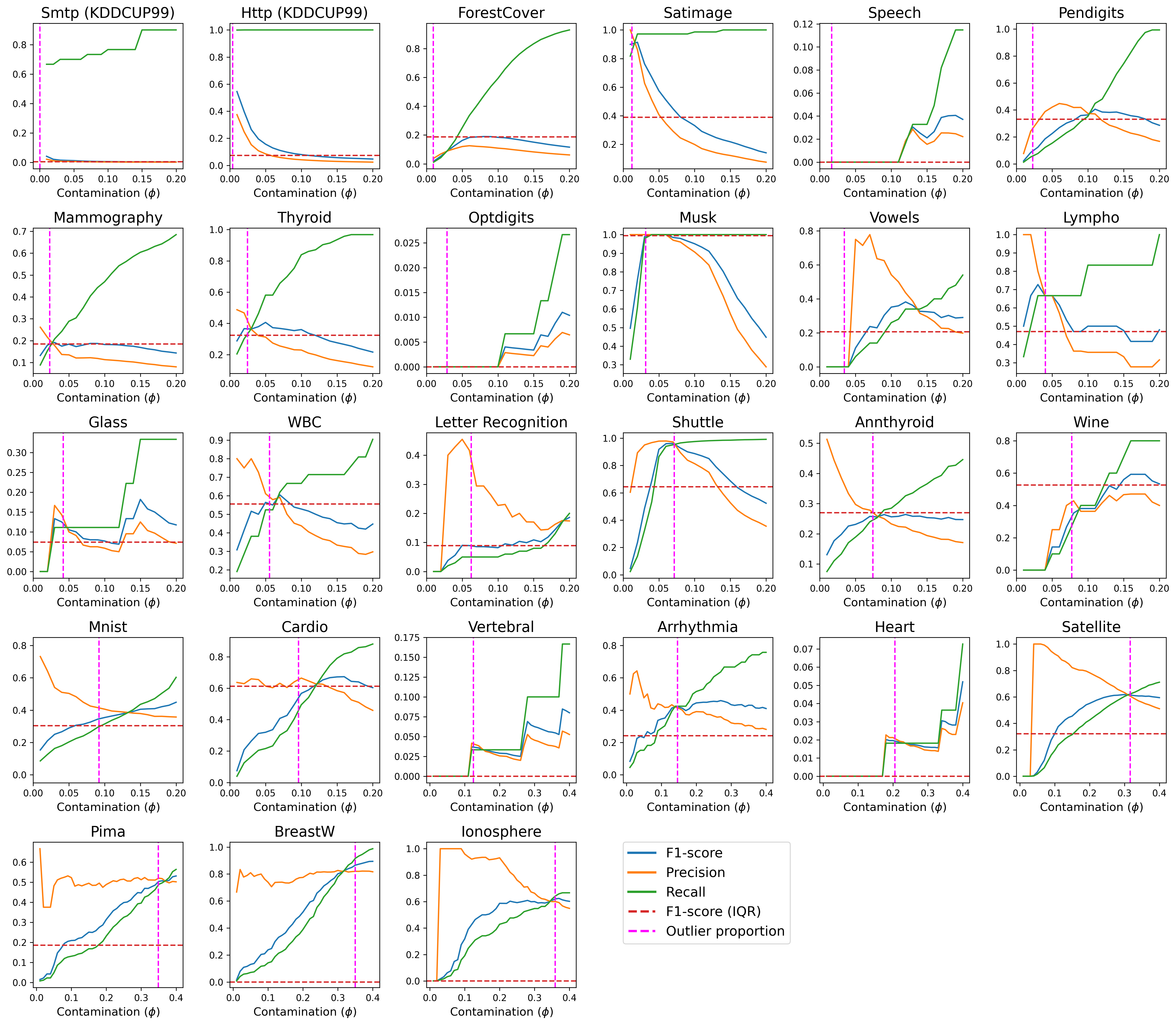}
	\caption{Sensitivity analysis for the contamination parameter $\phi$ using the 27 benchmark datasets.}
	\label{plot:sen}
\end{figure}

The results are illustrated in Figure~\ref{plot:sen}. Overall, performance is satisfactory when $\phi$ is chosen close to the true outlier proportion for each dataset. This indicates that, although the estimated proportion of outliers may not exactly match a given value of $\phi$, aligning $\phi$ with the true outlier proportion is a reasonable strategy.
Unfortunately, the true outlier proportion is generally unknown. While the IQR method is often satisfactory, it sometimes results in very poor performance, with zero F1-scores for some datasets. In contrast, using a fixed value of $\phi$ near $0.1$ often outperforms the IQR method, regardless of the actual outlier proportion in most cases (see the comparison between the blue solid and red dashed lines). Therefore, even when the true outlier proportion is unknown, we recommend using a default contamination parameter of $\phi = 0.1$ rather than relying on the IQR method. Nonetheless, the IQR method may still be preferred for certain philosophical reasons and generally performs reasonably well. We consider both approaches in our comparative analysis of OEDPM and other outlier detection methods.

\begin{table}[t!]
	\centering	
	\resizebox{0.8\width}{!}{%
		\begin{tabular}{llll}
			\toprule
			Year & Method                                                                          & Source       & $\phi$-type                               \\
			\midrule
			2000 & K-Nearest Neighbors (KNN) \cite{ramaswamy2000efficient}                                                      & PyOD         & Yes     \\
			2000 & Local Outlier Factor (LOF) \cite{42breunig2000lof}                                                     & PyOD         & Yes                  \\
			2001 & One-Class Support Vector Machines (OCSVM) \cite{scholkopf2001estimating}                                      & PyOD         & Yes            \\
			2003 & Principal Component Analysis (PCA) \cite{50shyu2003novel}                                             & PyOD         & Yes       \\
			2008 & Angle-Based Outlier Detector (ABOD)  \cite{48kriegel2008angle}                                           & PyOD         & Yes             \\
			2008 & Isolation Forest (IF) \cite{30liu2008isolation}                                                          & PyOD/scikit-learn        & Yes/No         \\
			2011 & Robust Trimmed Clustering (TCLUST)  \cite{garcia2011tclust}                                            & CRAN         & Yes                                        \\
			2014 & Autoencoder (AE) \cite{sakurada2014anomaly}                                                               & PyOD         & Yes                           \\
			2014 & Variational Autoencoder (VAE)  \cite{kingma2013auto}                                                 & PyOD         & Yes                                \\
			2016 & Contaminated Normal Mixtures   (ContaminatedMixt; CnMixt) \cite{punzo2016parsimonious} & CRAN         & No                                               \\
			2016 & Lightweight Online Detector of Anomalies (LODA)  \cite{pevny2016loda}                             & PyOD         & Yes                                    \\
			2018 & Deep One-Class Classification (DeepSVDD; DSVDD)   \cite{ruff2018deep}                                        & PyOD         & Yes                      \\
			2018 & Isolation using Nearest-Neighbor Ensembles (INNE) \cite{bandaragoda2018isolation}                              & PyOD         & Yes              \\
			2020 & Copula-Based Outlier Detection (COPOD)  \cite{li2020copod}                                        & PyOD         & Yes                               \\
			2020 & Rotation-Based Outlier Detection (ROD)  \cite{almardeny2020novel}                                        & PyOD         & Yes                  \\
			2021 & Neural Transformation Learning for Anomaly Detection (NeuTraL)  \cite{qiu2021neural}                & DeepOD       & No                 \\
			2021 & Internal Contrastive Learning (ICL) \cite{shenkar2021anomaly}                                            & DeepOD       & No                     \\
			2021 & Robust Collaborative Autoencoders (RCA) \cite{liu2021rca}                                        & DeepOD       & No             \\
			2022 & Empirical CDF-based Outlier Detection (ECOD) \cite{li2022ecod}                                   & PyOD         & Yes               \\
			2022 & Learnable Unified   Neighbourhood-Based Anomaly Ranking (LUNAR)  \cite{goodge2022lunar}               & PyOD         & Yes                 \\
			2023 & Deep Isolation Forest (DIF) \cite{xu2023deep}                                                    & PyOD/DeepOD         & Yes/No           \\
			2023 & Scale Learning-Based Deep   Anomaly Detection (SLAD) \cite{xu2023fascinating}                           & DeepOD       & No        \\  
			\bottomrule                             
		\end{tabular}
	}
	\caption{Details of the 22 competing methods, sorted in chronological order. Several methods require specifying a contamination-type parameter.}
	\label{tab:method}
\end{table}

\subsection{Comparison with other methods}

We now compare OEDPM with other methods for unsupervised outlier detection. We examine methodologies from the Python toolboxes PyOD (\url{https://pyod.readthedocs.io/en/latest/}) and DeepOD (\url{https://deepod.readthedocs.io/en/latest/}), as well as from two R packages for mixture-based methods available on CRAN (\url{https://cran.r-project.org/}).
In total, we evaluate 22 methods, ranging from classical approaches to state-of-the-art techniques. A summary of these methods is provided in Table~\ref{tab:method}, with abbreviations used consistently across Tables~\ref{tab:benchmark1}--\ref{tab:benchmark3}. Some methods require specifying a contamination-type parameter, which is typically either exactly or approximately equivalent to the outlier proportion in the detection results. Other methods do not have an option for specifying such parameters; instead, they automatically determine the outlier proportions based on their underlying rules. These methods are advantageous when there is no prior information about the true outlier proportion, as they do not require user-specified contamination parameters. Notably, OEDPM can operate in both ways: either by specifying $\phi$ or by using the IQR method.

\afterpage{
	\begin{landscape}
		\begin{table}[t!]
			\centering
			\resizebox{0.7\width}{!}{%
				\begin{tabular}{lrrrrrrrrrrrrrrrrrrr}
					\toprule
					& KNN            & LOF            & OCSVM & PCA   & ABOD           & IF             & TCLUST         & AE             & VAE            & LODA           & DSVDD       & INNE  & COPOD          & ROD            & ECOD           & LUNAR & DIF            & OEDPM          \\
					\midrule
					Smtp (KDDCUP99)    & 0.005          & 0.005          & 0.005 & 0.005 & -              & 0.005          & 0.005          & 0.005          & 0.004          & 0.023          & 0.004          & 0.005 & 0.004          & 0.004          & 0.004          & 0.004 & \textBF{0.094} & 0.005          \\
					Http (KDDCUP99)    & 0.004          & 0.002          & 0.075 & 0.075 & -              & 0.075          & 0.075          & 0.075          & 0.075          & 0.014          & 0.002          & 0.075 & 0.075          & 0.075          & 0.075          & 0.003 & \textBF{0.493} & 0.080          \\
					ForestCover        & 0.069          & 0.024          & 0.157 & 0.134 & 0.053          & 0.112          & 0.123          & 0.161          & 0.133          & 0.130          & 0.020          & 0.158 & 0.091          & 0.092          & 0.119          & -     & 0.132          & \textBF{0.184} \\
					Satimage           & 0.163          & 0.070          & 0.215 & 0.203 & 0.077          & 0.215          & 0.218          & 0.203          & 0.209          & 0.212          & 0.009          & 0.218 & 0.203          & 0.181          & 0.196          & 0.126 & \textBF{0.488} & 0.332          \\
					Speech             & 0.036          & 0.049          & 0.033 & 0.033 & \textBF{0.148} & 0.033          & 0.037          & 0.033          & 0.033          & 0.033          & 0.009          & 0.045 & 0.028          & -              & 0.033          & 0.033 & 0.012          & 0.000          \\
					Pendigits          & 0.090          & 0.073          & 0.247 & 0.240 & 0.074          & 0.225          & 0.102          & 0.242          & 0.242          & 0.311          & 0.024          & 0.171 & 0.202          & 0.275          & 0.244          & 0.083 & 0.229          & \textBF{0.363} \\
					Mammography        & 0.183          & 0.141          & 0.216 & 0.223 & -              & 0.218          & 0.218          & 0.226          & 0.226          & 0.173          & 0.074          & 0.220 & \textBF{0.286} & 0.125          & 0.284          & 0.171 & 0.015          & 0.182          \\
					Thyroid            & 0.374          & 0.095          & 0.344 & 0.323 & -              & 0.365          & \textBF{0.382} & 0.361          & 0.327          & 0.144          & 0.017          & 0.365 & 0.314          & 0.378          & 0.378          & 0.319 & 0.365          & 0.360          \\
					Optdigits          & 0.020          & 0.059          & 0.009 & 0.003 & 0.057          & 0.057          & 0.003          & 0.003          & 0.003          & 0.030          & \textBF{0.182} & 0.015 & 0.036          & 0.083          & 0.033          & 0.027 & 0.026          & 0.000          \\
					Musk               & 0.183          & 0.201          & 0.480 & 0.480 & 0.003          & 0.480          & 0.480          & 0.480          & 0.480          & 0.480          & 0.059          & 0.480 & 0.356          & -              & 0.381          & 0.035 & 0.611          & \textBF{0.951} \\
					Vowels             & \textBF{0.566} & 0.358          & 0.214 & 0.133 & 0.427          & 0.163          & 0.480          & 0.133          & 0.143          & 0.143          & 0.000          & 0.276 & 0.031          & 0.102          & 0.122          & 0.469 & 0.000          & 0.351          \\
					Lympho             & 0.381          & 0.381          & 0.476 & 0.476 & 0.231          & \textBF{0.571} & 0.381          & \textBF{0.571} & 0.546          & 0.381          & 0.286          & 0.381 & \textBF{0.571} & 0.476          & \textBF{0.571} & 0.476 & 0.000          & 0.500          \\
					Glass              & 0.069          & \textBF{0.138} & 0.065 & 0.065 & \textBF{0.138} & 0.129          & 0.065          & 0.065          & 0.067          & 0.065          & 0.065          & 0.129 & 0.065          & 0.129          & 0.129          & 0.129 & 0.000          & 0.077          \\
					WBC                & 0.500          & 0.500          & 0.509 & 0.509 & 0.371          & 0.509          & 0.475          & 0.441          & 0.367          & 0.441          & 0.102          & 0.475 & 0.509          & 0.509          & 0.441          & 0.475 & 0.000          & \textBF{0.528} \\
					Letter Recognition & 0.409          & 0.451          & 0.162 & 0.100 & 0.392          & 0.123          & \textBF{0.500} & 0.100          & 0.100          & 0.115          & 0.069          & 0.231 & 0.085          & 0.077          & 0.085          & 0.392 & 0.037          & 0.082          \\
					Shuttle            & 0.214          & 0.127          & 0.812 & 0.803 & 0.187          & 0.820          & 0.834          & 0.810          & 0.803          & 0.008          & 0.243          & 0.824 & 0.820          & 0.794          & 0.818          & 0.194 & 0.097          & \textBF{0.887} \\
					Annthyroid         & 0.313          & 0.231          & 0.252 & 0.241 & -              & 0.325          & \textBF{0.493} & 0.268          & 0.241          & 0.137          & 0.249          & 0.271 & 0.242          & 0.346          & 0.303          & 0.236 & 0.165          & 0.256          \\
					Wine               & 0.000          & 0.182          & 0.087 & 0.348 & 0.000          & 0.087          & 0.000          & 0.174          & 0.333          & \textBF{0.522} & 0.174          & 0.000 & \textBF{0.522} & 0.435          & 0.261          & 0.000 & 0.000          & 0.381          \\
					Mnist              & \textBF{0.425} & 0.292          & 0.404 & 0.394 & 0.338          & 0.315          & 0.400          & 0.394          & 0.394          & 0.183          & 0.290          & 0.411 & 0.249          & 0.162          & 0.189          & 0.364 & 0.408          & 0.354          \\
					Cardio             & 0.310          & 0.176          & 0.513 & 0.613 & -              & 0.529          & 0.256          & \textBF{0.618} & 0.613          & 0.596          & 0.128          & 0.384 & 0.535          & 0.596          & 0.518          & 0.351 & 0.323          & 0.567          \\
					Vertebral          & 0.040          & 0.040          & 0.037 & 0.000 & 0.037          & 0.037          & 0.037          & 0.000          & 0.000          & 0.037          & 0.037          & 0.037 & 0.000          & \textBF{0.074} & 0.037          & 0.037 & 0.000          & 0.000          \\
					Arrhythmia         & 0.342          & 0.252          & 0.339 & 0.357 & 0.347          & 0.446          & 0.339          & 0.357          & 0.357          & 0.357          & 0.268          & 0.286 & 0.429          & -              & \textBF{0.464} & 0.268 & 0.173          & 0.346          \\
					Heart              & 0.000          & 0.027          & 0.000 & 0.000 & 0.000          & 0.000          & 0.024          & 0.000          & 0.000          & 0.024          & \textBF{0.122} & 0.024 & 0.000          & 0.000          & 0.000          & 0.000 & 0.000          & 0.000          \\
					Satellite          & 0.293          & 0.206          & 0.476 & 0.454 & 0.256          & 0.463          & 0.352          & 0.455          & \textBF{0.481} & 0.431          & 0.196          & 0.399 & 0.405          & 0.333          & 0.390          & 0.303 & 0.009          & 0.321          \\
					Pima               & 0.199          & 0.150          & 0.226 & 0.232 & 0.253          & 0.273          & 0.220          & 0.226          & 0.231          & 0.209          & 0.151          & 0.209 & 0.284          & \textBF{0.290} & 0.238          & 0.244 & 0.015          & 0.209          \\
					BreastW            & 0.364          & 0.000          & 0.422 & 0.448 & -              & 0.442          & 0.390          & 0.429          & \textBF{0.458} & 0.448          & 0.208          & 0.130 & 0.448          & 0.448          & 0.448          & 0.429 & 0.000          & 0.239          \\
					Ionosphere         & 0.415          & 0.377          & 0.435 & 0.410 & \textBF{0.461} & 0.435          & 0.420          & 0.410          & 0.410          & 0.422          & 0.261          & 0.435 & 0.385          & 0.398          & 0.407          & 0.435 & 0.016          & 0.318          \\
					\midrule
					Average            & 0.221          & 0.171          & 0.267 & 0.270 & 0.143          & 0.276          & 0.271          & 0.268          & 0.269          & 0.225          & 0.120          & 0.246 & 0.266          & 0.236          & 0.266          & 0.207 & 0.137          & \textBF{0.292}\\
					\bottomrule
				\end{tabular}
			}
			\caption{F1-scores for outlier detection with the contamination-type parameter set to $\phi=0.1$. Empty cells indicate that the method was not executed for the respective benchmark dataset. The last row represents the average F1-scores across all benchmark datasets. The maximum value in each row is bolded to emphasize the best performance.}
			\label{tab:benchmark1}
		\end{table}

		\begin{table}[t!]
			\centering
			\resizebox{0.7\width}{!}{%
				\begin{tabular}{lrrrrrrrrrrrrrrrrrrr}
					\toprule
					& KNN            & LOF            & OCSVM & PCA   & ABOD           & IF             & TCLUST         & AE             & VAE            & LODA           & DSVDD       & INNE  & COPOD          & ROD            & ECOD           & LUNAR & DIF            & OEDPM          \\
					\midrule
					Smtp (KDDCUP99)    & 0.003          & 0.003          & 0.002 & 0.002          & -              & 0.002          & 0.003          & 0.002 & 0.002          & 0.014 & 0.003          & 0.002          & 0.002          & 0.002          & 0.002          & 0.002          & \textBF{0.070} & 0.003          \\
					Http (KDDCUP99)    & 0.002          & 0.001          & 0.038 & 0.038          & -              & 0.038          & 0.038          & 0.038 & 0.038          & 0.002 & 0.001          & 0.038          & 0.038          & 0.038          & 0.038          & 0.002          & \textBF{0.431} & 0.047          \\
					ForestCover        & 0.055          & 0.022          & 0.091 & 0.089          & 0.045          & 0.081          & 0.081          & 0.089 & 0.089          & 0.087 & 0.001          & 0.090          & 0.071          & 0.078          & 0.083          & 0.053          & \textBF{0.315} & 0.116          \\
					Satimage           & 0.108          & 0.044          & 0.115 & 0.112          & 0.059          & 0.114          & 0.115          & 0.114 & 0.112          & 0.112 & 0.016          & 0.115          & 0.112          & 0.101          & 0.107          & 0.094          & \textBF{0.282} & 0.141          \\
					Speech             & 0.034          & 0.042          & 0.028 & 0.028          & \textBF{0.107} & 0.033          & 0.025          & 0.028 & 0.028          & 0.035 & 0.030          & 0.035          & 0.033          & -              & 0.033          & 0.033          & 0.034          & 0.037          \\
					Pendigits          & 0.102          & 0.059          & 0.188 & 0.197          & 0.066          & 0.203          & 0.129          & 0.199 & 0.199          & 0.197 & 0.060          & 0.143          & 0.174          & 0.199          & 0.183          & 0.080          & 0.263          & \textBF{0.286} \\
					Mammography        & 0.142          & 0.110          & 0.165 & 0.176          & -              & 0.163          & 0.160          & 0.174 & 0.165          & 0.154 & 0.038          & 0.147          & \textBF{0.178} & 0.140          & \textBF{0.178} & 0.143          & 0.091          & 0.143          \\
					Thyroid            & 0.229          & 0.097          & 0.212 & 0.212          & -              & 0.215          & 0.219          & 0.212 & 0.215          & 0.085 & 0.028          & 0.215          & 0.215          & 0.217          & 0.219          & 0.205          & \textBF{0.430} & 0.216          \\
					Optdigits          & 0.025          & 0.077          & 0.022 & 0.015          & 0.053          & 0.099          & 0.034          & 0.015 & 0.015          & 0.015 & 0.040          & 0.032          & 0.070          & \textBF{0.102} & 0.047          & 0.042          & 0.031          & 0.010          \\
					Musk               & 0.166          & 0.115          & 0.273 & 0.273          & 0.007          & 0.273          & 0.273          & 0.273 & 0.273          & 0.273 & 0.251          & 0.273          & 0.268          & -              & 0.265          & 0.082          & 0.413          & \textBF{0.448} \\
					Vowels             & \textBF{0.359} & 0.296          & 0.158 & 0.094          & 0.274          & 0.123          & 0.287          & 0.105 & 0.106          & 0.153 & 0.065          & 0.223          & 0.059          & 0.106          & 0.088          & 0.282          & 0.103          & 0.290          \\
					Lympho             & 0.333          & 0.364          & 0.333 & 0.333          & 0.195          & 0.333          & 0.278          & 0.324 & 0.333          & 0.278 & 0.167          & 0.333          & 0.333          & 0.333          & 0.333          & 0.333          & 0.000          & \textBF{0.480} \\
					Glass              & 0.174          & \textBF{0.240} & 0.115 & 0.115          & 0.151          & 0.115          & 0.115          & 0.115 & 0.115          & 0.154 & 0.039          & 0.115          & 0.039          & 0.115          & 0.115          & 0.192          & 0.000          & 0.118          \\
					WBC                & 0.418          & 0.409          & 0.392 & 0.351          & 0.350          & 0.392          & 0.412          & 0.379 & 0.351          & 0.392 & 0.041          & 0.351          & 0.392          & 0.351          & 0.330          & 0.412          & 0.103          & \textBF{0.447} \\
					Letter Recognition & \textBF{0.388} & 0.382          & 0.176 & 0.114          & 0.357          & 0.157          & 0.329          & 0.114 & 0.119          & 0.119 & 0.119          & 0.205          & 0.095          & 0.110          & 0.129          & 0.324          & 0.064          & 0.186          \\
					Shuttle            & 0.201          & 0.128          & 0.518 & 0.518          & 0.181          & 0.524          & \textBF{0.527} & 0.518 & 0.519          & 0.343 & 0.127          & 0.522          & 0.522          & 0.517          & 0.523          & 0.183          & 0.225          & 0.524          \\
					Annthyroid         & 0.310          & 0.248          & 0.216 & 0.213          & -              & 0.320          & \textBF{0.414} & 0.215 & 0.231          & 0.060 & 0.295          & 0.243          & 0.247          & 0.330          & 0.280          & 0.242          & 0.186          & 0.247          \\
					Wine               & 0.000          & 0.412          & 0.222 & 0.389          & 0.000          & 0.333          & 0.056          & 0.389 & 0.389          & 0.389 & 0.056          & 0.389          & 0.444          & 0.444          & 0.333          & 0.056          & 0.000          & \textBF{0.533} \\
					Mnist              & 0.411          & 0.280          & 0.430 & 0.422          & 0.314          & 0.334          & 0.385          & 0.422 & 0.423          & 0.344 & 0.148          & 0.428          & 0.315          & 0.204          & 0.288          & 0.357          & 0.411          & \textBF{0.449} \\
					Cardio             & 0.320          & 0.188          & 0.554 & 0.579          & -              & 0.513          & 0.413          & 0.579 & 0.579          & 0.369 & 0.162          & 0.550          & 0.528          & 0.594          & 0.565          & 0.277          & 0.500          & \textBF{0.603} \\
					Vertebral          & 0.028          & 0.080          & 0.051 & \textBF{0.103} & 0.071          & 0.077          & 0.026          & 0.100 & \textBF{0.103} & 0.000 & 0.051          & 0.051          & 0.000          & \textBF{0.103} & \textBF{0.103} & 0.026          & 0.000          & 0.029          \\
					Arrhythmia         & 0.436          & 0.427          & 0.446 & 0.433          & 0.410          & 0.497          & 0.471          & 0.433 & 0.446          & 0.408 & 0.280          & 0.395          & 0.484          & \textBF{0.510} & 0.484          & 0.459          & 0.377          & 0.444          \\
					Heart              & 0.019          & 0.019          & 0.018 & 0.018          & 0.036          & 0.018          & 0.018          & 0.019 & 0.018          & 0.055 & \textBF{0.110} & 0.037          & 0.018          & 0.018          & 0.037          & 0.018          & 0.000          & 0.020          \\
					Satellite          & 0.436          & 0.302          & 0.565 & 0.530          & 0.347          & \textBF{0.616} & 0.603          & 0.576 & 0.536          & 0.506 & 0.309          & \textBF{0.616} & 0.491          & 0.347          & 0.458          & 0.429          & 0.477          & 0.551          \\
					Pima               & 0.371          & 0.303          & 0.379 & 0.374          & 0.410          & 0.408          & 0.360          & 0.388 & 0.379          & 0.289 & 0.280          & 0.365          & 0.431          & \textBF{0.441} & 0.389          & 0.422          & 0.147          & 0.314          \\
					BreastW            & 0.631          & 0.033          & 0.692 & 0.718          & -              & 0.713          & 0.638          & 0.658 & 0.707          & 0.723 & 0.447          & 0.266          & 0.723          & \textBF{0.729} & 0.723          & 0.692          & 0.000          & 0.518          \\
					Ionosphere         & \textBF{0.714} & 0.600          & 0.684 & 0.602          & 0.711          & 0.653          & 0.660          & 0.605 & 0.582          & 0.643 & 0.449          & 0.704          & 0.477          & 0.480          & 0.487          & \textBF{0.714} & 0.274          & 0.587          \\
					\midrule
					Average      & 0.238 &	0.195 &	0.262 &	0.261 &	0.153 &	0.272 &	0.262 &	0.262 &	0.262 &	0.230 &	0.134 &	0.255 &	0.250 &	0.245 &	0.253 &	0.228 &	0.194 &	\textBF{0.288} \\
					\bottomrule
				\end{tabular}
			}
			\caption{F1-scores for outlier detection with the contamination-type parameter set to $\phi=0.2$. Empty cells indicate that the method was not executed for the respective benchmark dataset. The last row represents the average F1-scores across all benchmark datasets. The maximum value in each row is bolded to emphasize the best performance.}
			\label{tab:benchmark2}
		\end{table}
	\end{landscape}
}

We apply all competing methods, including OEDPM, to detect outliers in the 27 benchmark datasets listed in Table~\ref{tab:data}. For OEDPM, we use $M=100$ ensemble components, which typically provide a sufficient ensemble size to minimize potential bias. For methods requiring a contamination-type parameter (including OEDPM with $\phi$), we test two values: $\phi=0.1$ and $\phi=0.2$. Additionally, OEDPM is evaluated using the IQR method to compare with methods that do not use contamination-type parameters. We calculate F1-scores and runtimes for each method across the benchmark datasets. To ensure a fair comparison, we do not directly compare methods with and without a contamination-type parameter.

Table~\ref{tab:benchmark1} and Table~\ref{tab:benchmark2} show the F1-scores for methods with $\phi=0.1$ and $\phi=0.2$, respectively. These results indicate that, while not always the case, recent methods generally perform better than classical methods in outlier detection. Among the state-of-the-art methods, our proposed OEDPM performs exceptionally well, often outperforming other competitors in terms of F1-scores. Specifically, in Table~\ref{tab:benchmark1} with $\phi=0.1$, OEDPM leads in five benchmark datasets, surpassing other methods. Similarly, for $\phi=0.2$, OEDPM wins in seven benchmark datasets, as shown in Table~\ref{tab:benchmark2}.
Even when OEDPM does not secure the top position, its F1-scores are generally competitive. The final rows of both tables show the average F1-scores across all benchmark datasets, with OEDPM demonstrating the highest average F1-score, confirming its superior performance.

Table~\ref{tab:benchmark3} presents the F1-scores for methods that do not require a contamination-type parameter, including OEDPM with the IQR method. OEDPM clearly performs very well across most benchmark datasets. Notably, despite its straightforward construction, OEDPM often outperforms more complex recent methods, including those based on neural networks.

\begin{table}[t!]
	\centering
	\resizebox{0.8\width}{!}{%
\begin{tabular}{lrrrrrrrr}
	\toprule
	& IF (w/o $\phi$)      & CnMixt         & NeuTraL          & ICL    & RCA         & DIF (w/o $\phi$)     & SLAD           & OEDPM (IQR)    \\
	\midrule
Smtp (KDDCUP99)    & 0.003                     & \textBF{0.012} & 0.005          & 0.000 & 0.005          & 0.005          & 0.005          & 0.004          \\
Http (KDDCUP99)    & 0.062                     & 0.000          & -              & -     & \textBF{0.075} & 0.075          & 0.075          & 0.074          \\
ForestCover        & 0.047                     & 0.043          & -              & -     & -              & 0.153          & -              & \textBF{0.186} \\
Satimage           & 0.185                     & 0.026          & 0.022          & 0.000 & 0.212          & 0.215          & 0.190          & \textBF{0.390} \\
Speech             & 0.000                     & -              & 0.009          & 0.023 & 0.032          & \textBF{0.033} & 0.019          & 0.000          \\
Pendigits          & 0.102                     & 0.000          & 0.043          & 0.038 & 0.192          & 0.247          & 0.114          & \textBF{0.331} \\
Mammography        & 0.188                     & 0.000          & 0.075          & 0.038 & \textBF{0.202} & 0.193          & 0.060          & 0.185          \\
Thyroid            & \textBF{0.363}            & 0.000          & 0.115          & 0.093 & 0.345          & 0.344          & 0.136          & 0.324          \\
Optdigits          & \textBF{0.085}            & -              & 0.083          & 0.036 & 0.006          & 0.003          & 0.018          & 0.000          \\
Musk               & 0.425                     & 0.956          & 0.480          & 0.356 & 0.472          & 0.480          & 0.223          & \textBF{0.995} \\
Vowels             & 0.145                     & 0.154          & 0.327          & 0.143 & 0.306          & 0.214          & \textBF{0.347} & 0.207          \\
Lympho             & 0.200                     & 0.000          & 0.095          & 0.000 & 0.455          & 0.381          & 0.381          & \textBF{0.471} \\
Glass              & 0.065                     & 0.000          & \textBF{0.194} & 0.065 & \textBF{0.194} & 0.129          & \textBF{0.194} & 0.074          \\
WBC                & \multicolumn{1}{r}{0.426} & 0.144          & 0.034          & 0.034 & 0.517          & 0.475          & 0.203          & \textBF{0.556} \\
Letter Recognition & 0.096                     & 0.000          & 0.369          & 0.246 & 0.234          & 0.154          & \textBF{0.377} & 0.089          \\
Shuttle            & 0.730                     & -              & 0.274          & 0.029 & 0.814          & 0.759          & \textBF{0.834} & 0.645          \\
Annthyroid         & \textBF{0.293}            & 0.000          & 0.177          & 0.061 & 0.262          & 0.257          & 0.131          & 0.270          \\
Wine               & 0.167                     & 0.000          & 0.087          & 0.435 & 0.000          & 0.000          & 0.000          & \textBF{0.526} \\
Mnist              & 0.318                     & -              & 0.225          & 0.163 & \textBF{0.394} & 0.368          & -              & 0.304          \\
Cardio             & 0.513                     & 0.583          & 0.106          & 0.050 & 0.378          & 0.446          & 0.184          & \textBF{0.612} \\
Vertebral          & 0.037                     & 0.000          & \textBF{0.111} & 0.000 & 0.039          & 0.037          & 0.037          & 0.000          \\
Arrhythmia         & 0.154                     & -              & 0.196          & 0.089 & 0.319          & \textBF{0.357} & 0.214          & 0.241          \\
Heart              & 0.000                     & \textBF{0.140} & 0.000          & 0.024 & 0.000          & 0.000          & 0.073          & 0.000          \\
Satellite          & \textBF{0.531}            & 0.406          & 0.240          & 0.289 & 0.405          & 0.377          & 0.275          & 0.321          \\
Pima               & 0.266                     & \textBF{0.427} & 0.197          & 0.174 & 0.212          & 0.232          & 0.145          & 0.185          \\
BreastW            & \textBF{0.931}            & 0.898          & 0.201          & 0.351 & 0.432          & 0.409          & 0.396          & 0.000          \\
Ionosphere         & 0.667                     & \textBF{0.703} & 0.286          & 0.323 & 0.444          & 0.435          & 0.236          & 0.000          \\
\midrule
Average            & \textBF{0.259}            & 0.166          & 0.146          & 0.113 & 0.257          & 0.251          & 0.180          &	\textBF{0.259} \\
	\bottomrule       
\end{tabular}
	}
	\caption{F1-scores for outlier detection without a contamination-type parameter. Empty cells indicate that the method was not executed for the respective benchmark dataset. 	The last row represents the average F1-scores across all benchmark datasets. The maximum value in each row is bolded to emphasize the best performance.
}
	\label{tab:benchmark3}
\end{table}

\begin{figure}[t!]
	\centering
	\includegraphics[width = 13cm]{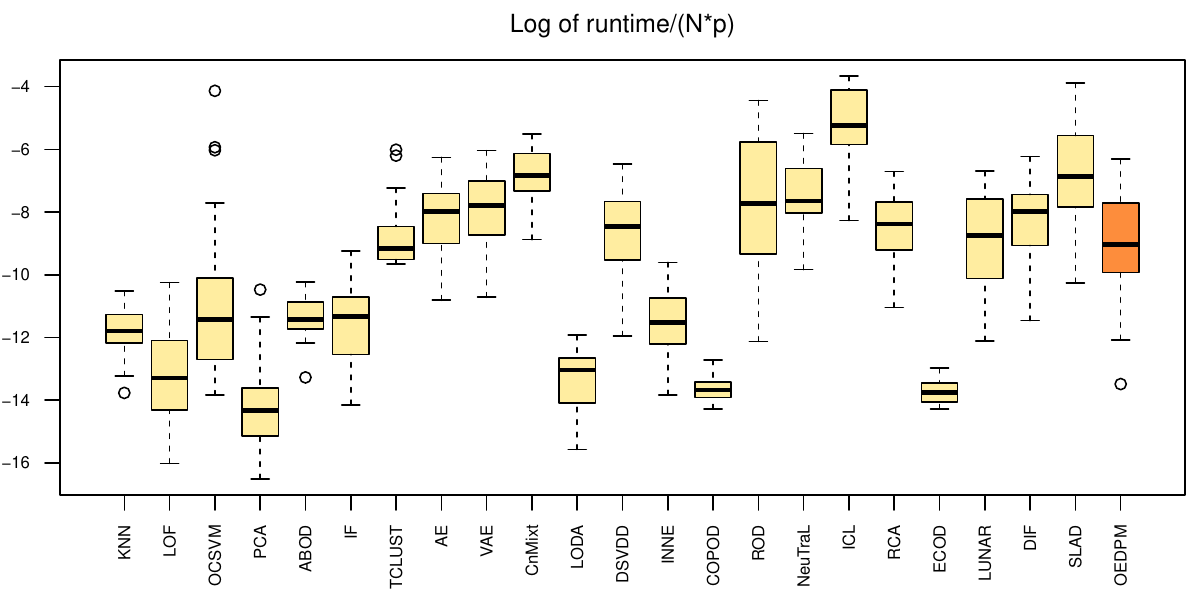}
	\caption{Logarithm of runtime (seconds) divided by data size ($N\times p$) for the benchmark datasets.}
	\label{plot:runtime}
\end{figure}

Figure~\ref{plot:runtime} compares the runtime of OEDPM with the runtimes of competing methods. We calculated the logarithm of the ratio of runtime (in seconds) to data size ($N \times p$) for the results in Tables~\ref{tab:benchmark1}--\ref{tab:benchmark3}. 
The comparison reveals that recent methods generally have longer runtimes than classical methods, often due to their reliance on computationally intensive structures such as neural networks. In contrast, OEDPM demonstrates a reasonable runtime while maintaining excellent performance. This efficiency is largely due to the use of variational inference and ensemble analysis, as detailed in Section~\ref{sec2}.

\section{Discussion}
\label{sec7}

This study introduced OEDPM for unsupervised outlier detection. By integrating two outlier ensemble techniques into the DPGM with variational inference, OEDPM provides unique advantages not achievable with traditional Gaussian mixture modeling. Specifically, the subspace ensemble with random projection facilitates efficient data characterization through dimensionality reduction. This approach makes the data suitable for Gaussian modeling, even when they significantly deviate from Gaussian distributions. Additionally, the subsampling ensemble addresses the challenge of long computation times--a major issue in mixture modeling--without compromising detection accuracy. Our numerical analyses confirm the effectiveness of OEDPM.

A key factor in the success of OEDPM is the outlier ensemble with random projection, which involves linear projection onto smaller subspaces. While this linear approach contributes to simplicity and robustness, it may also be viewed as a limitation in the modeling process. Future research should explore alternative methods, such as nonlinear projection, to enhance the construction of outlier ensembles.

\section*{Acknowledgment}
Dongwook Kim and Juyeon Park contributed equally to this work. The research was supported by the Yonsei University Research Fund of 2021-22-0032 and by the National Research Foundation of Korea (NRF) grant funded by the Korean government (MSIT) (2022R1C1C1006735, RS-2023-00217705).

\bibliographystyle{chicago}
\bibliography{Ref}

\end{document}